\documentclass[pdflatex,sn-mathphys-num, iicol]{sn-jnl}

\usepackage{graphicx}%
\usepackage{multirow}%
\usepackage{amsmath,amssymb,amsfonts}%
\usepackage{amsthm}%
\usepackage{mathrsfs}%
\usepackage[title]{appendix}%
\usepackage{xcolor}%
\usepackage{textcomp}%
\usepackage{manyfoot}%
\usepackage{booktabs}%
\usepackage{bm}
\usepackage{algorithm}%
\usepackage{algorithmicx}%
\usepackage{algpseudocode}%
\usepackage{listings}%
\usepackage{colortbl}
\usepackage{makecell}
\usepackage{subcaption}
\usepackage{balance}
\usepackage{xurl}

\theoremstyle{thmstyleone}%
\theoremstyle{thmstyletwo}%

\theoremstyle{thmstylethree}%
\begin{document}

\title[ACTrack]{
    Models as Tools: An Agentic Coordination Framework for Unified Multimodal Visual Tracking
}


\author[1,2]{\fnm{Wenrui} \sur{Cai}}\equalcont{Equal Contribution}
\author[1,2]{\fnm{Yuzhe} \sur{Li}}\equalcont{Equal Contribution}
\author*[1,2,3]{\fnm{Qingjie} \sur{Liu}}\email{qingjie.liu@buaa.edu.cn}
\author[1,2,3]{\fnm{Yunhong} \sur{Wang}}

\affil[1]{\orgdiv{State Key Laboratory of Virtual Reality Technology and System}, \orgname{Beihang University}}
\affil[2]{\orgdiv{School of Computer Science and Engineering}, \orgname{Beihang University}}
\affil[3]{\orgdiv{Hangzhou Innovation Institute}, \orgname{Beihang University}}


\abstract{

    Most current visual trackers adopt a matching-based one-stream Transformer architecture trained exclusively on visual tracking datasets, 
    whose performance gains depend heavily on the length of the input context, and have now reached a bottleneck.
    While high-performance tracking increasingly relies on foundation models, existing methods use them monolithically, adapting a foundation model into a tracker or modify a segmentation foundation model into a tracking pipeline, which fails to exploit complementary strengths. 
    Matching-based trackers excel at instance-level correspondence but lack semantic discrimination and fine-grained foreground perception, whereas segmentation foundation models produce precise masks yet struggle with instance discrimination and multimodal extension. 
    Both paradigms also lack error-correction capabilities for long-term tracking.
    To address these issues, we propose ACTrack, an agentic coordination framework that treats heterogeneous models as invocable tools under an event-triggered mechanism. 
    ACTrack coordinates a Tracker-based Instance Matching Tool for target discrimination, 
    a SAM3 Motion Tool for mask-derived motion priors, 
    a SAM3 Perception Tool for detecting distractors and instance-conflict cues, 
    and a VLM Reprompt Tool activated only under persistent conflict to mitigate error accumulation. 
    We design a complete tool-invocation trigger mechanism and an inter-tool coordination mechanism, enabling the complementary strengths of different model tools to be fully integrated.
    Experiments show that ACTrack substantially surpasses the strongest and the largest trackers on eight RGB benchmarks including LaSOT, GOT-10K, and TrackingNet. 
    Furthermore, a parameter-efficient adaptation strategy enables parameter sharing and reuse across tools, 
    achieving unified multimodal tracking with only 30\% trainable parameters while substantially outperforming prior methods on multimodal benchmarks such as LasHeR, VisEvent, TNL2K, and DepthTrack.
}




\maketitle

\section{Introduction}
\label{sec:intro}

Visual object tracking aims to localize an arbitrary target throughout a video sequence from its initial annotation. 
Recent trackers have largely followed a matching-based one-stream Transformer paradigm, where template and search regions are jointly encoded and the target is localized through discriminative correspondence~\cite{ye_2022_joint,Cui_2022_MixFormer}. 
This paradigm has produced strong progress through sequence-level prediction and autoregressive localization~\cite{Chen_2023_CVPR_seqtrack,Wei_2023_CVPR_autoregressive}, online target queries and updated auxiliary templates~\cite{ODTrack,Bai_2024_CVPR_ARTrackV2,Xie_2024_CVPR_AQATrack}, and richer video-level context propagation~\cite{Cai_2024_CVPR_HIPTrack,kang2025exploring_mcitrack,li2025mambalct}. 

However, robust tracking requires more than matching, a tracker is supposed to preserve the annotated instance, perceive fine foreground boundaries, recover from target disappearance, reject similar distractors, and adapt to multimodal inputs. 
When all these capabilities are compressed into a single prediction stream trained mainly on tracking data, further gains in performance are prone to hitting a fundamental bottleneck, and become increasingly dependent on context length and model scale~\cite{ODTrack,kang2025exploring_mcitrack,Wu_2026_CVPR_token_compression}.

Foundation models provide a natural route beyond this bottleneck. Large pretrained visual encoders offer transferable representations~\cite{oquab2024dinov2}, and parameter-efficient tuning makes it practical to adapt such backbones to tracking without full fine-tuning~\cite{LoRAT,lin2025lorat_v2}. Recent scalable trackers further show the benefit of lightweight adaptation and spatio-temporal tuning~\cite{Cai_2025_CVPR_SPMTrack}. 
Meanwhile, SAM-style models introduce promptable mask prediction, context memory, and fine-grained foreground perception~\cite{ravi2025sam2}; SAM3 further strengthens concept-level segmentation and general visual perception~\cite{carion2026sam3segmentconcepts}. Recent SAM-based trackers show that segmentation models can provide useful motion and mask priors~\cite{yang2026samurai} and distractor-aware memory~\cite{Videnovic_2025_CVPR_sam2.1++}. 
However, most existing uses of foundation models remain monolithic: a foundation model is converted into a matching tracker, or a segmentation model is modified into a complete tracking pipeline. Such designs obscure the fact that different models are trained for different objectives and therefore fail in different ways.

This dilemma raises a central question: \emph{Can visual tracking be improved by coordinating different models as tools, rather than forcing their capabilities into one tracker?} In this paper, we observe that matching-based trackers are effective at instance-level correspondence across frames, but they can drift to similar objects and often lack semantic discrimination and fine-grained foreground awareness. SAM3 provides strong mask-level perception and motion priors, but may follow a semantically plausible object extent instead of the annotated instance, and its memory can be contaminated by incorrect results, leading to error accumulation. 
Vision-language model provides high-level semantic understanding~\cite{bai2025qwen3vltechnicalreport,sun2024chattracker,yu2024merlin,wang2026vptrackerglobalvisionlanguagetracking,huang2026adaptivebatchwisesamplescheduling,huang2026doesreasoningmodelimplicitly,huang2026realtimealignedrewardmodel}, making it well suited for judging and arbitrating among ambiguous candidate hypotheses; however, using VLMs densely for frame-by-frame localization is computationally expensive and unstable for localization. 
These complementary strengths and weaknesses suggest that the key is not to replace the tracker with another model, but to decide when each model should provide the evidence it excels at, and to enable mutual error correction and verification.

Based on the observation, we propose \textbf{ACTrack}, an agentic coordination framework that treats models as invocable tools for visual tracking. 
The agentic nature of ACTrack lies in its explicit tool coordination, conflict-aware arbitration, and event-triggered intervention: instead of asking every model to process every frame, ACTrack lets efficient visual model tools handle routine localization, monitors disagreement among tools, and invokes high-level semantic reasoning only when persistent conflict suggests that the current evidence has become unreliable. 
ACTrack coordinates a Tracker-based \emph{Instance Matching Tool}, a SAM3-based \emph{Motion Tool}, a SAM3-based \emph{Perception Tool}, and a VLM-based \emph{Reprompt Tool}. The instance matching tool provides default bounding box localization of the target. The motion tool generates mask-derived motion priors that re-anchor the search region, while the perception tool detects similar instances within the search region and determines whether the motion tool and the instance matching tool are in conflict. 
Persistent conflicts indicate that cross-tool coordination has become unreliable; therefore, the VLM Reprompt Tool is invoked to determine which tool output is more reliable and to refresh the states of another tool.

Beyond the application to RGB tracking, the same tool-coordination framework in ACTrack can also support multimodal tracking.
Existing unified multimodal trackers aim to cover RGB, RGB-D, RGB-T, RGB-Event, and RGB-Language settings through a single architecture~\cite{chen2023unified_seqtrackv2,sutrack,cai2026unimdtracklearningdecoupledmemory}. ACTrack is complementary to these efforts. 
Instead of requiring SAM3 to natively process every modality, 
We further introduce ACTrack-EM, which adapts the SAM3 visual backbone into a unified multimodal tracker through parameter-efficient fine-tuning, while keeping the cross-tool coordination procedure unchanged. This design decouples modality adaptation from tool coordination: the Instance Matching Tool handles heterogeneous RGB and RGB-X (D/T/E) inputs, whereas the remaining tools continue to operate on the RGB stream.
With the parameter-efficient fine-tuning design, ACTrack-EM substantially reduces the overall parameter count, supports unified tracking across all modalities, and fully preserves the coordination advantages of ACTrack.
Our experiments show that, in both RGB and multimodal tracking, ACTrack and ACTrack-EM substantially outperform all previous state-of-the-art trackers across all model scales.

Our contributions are summarized as follows. 
\textbf{(1)} We analyze the strengths and failure modes of heterogeneous models as tracking tools, and formulate visual tracking as \emph{agentic tool coordination} rather than a monolithic tracker. 
\textbf{(2)} We propose ACTrack, a tool-coordinated tracking framework in which matching-based tracking, SAM3-based motion and perception, and VLM reprompting interact through search-region re-anchoring, instance-conflict detection, and prompt renewal. 
\textbf{(3)} We develop a parameter-efficient unified multimodal version ACTrack-EM, enabling tool sharing and reuse across RGB and RGB-X tracking while substantially reducing additional parameters. 
\textbf{(4)} Extensive experiments show that ACTrack achieves highly competitive performance across standard RGB and multimodal benchmarks, demonstrating the effectiveness of coordinated tools over isolated model scaling.

\section{Related Work}
\label{sec:related}

\subsection{Matching-Based Visual Tracking}

Most modern visual trackers are built on a matching principle: the target specified in the first frame is represented by a template, and later frames are searched by comparing this template with candidate regions. Siamese trackers make this principle explicit~\cite{bertinetto2016fully,li2018high}. 

One-stream Transformer trackers generalize template-search matching from local correlation to global attention-based relation modeling. TransT uses self-attention and cross-attention to fuse template and search features~\cite{chen2021transformer}, and STARK models spatial-temporal dependencies with a Transformer encoder-decoder~\cite{yan2021learning}. 
One-stream trackers further merge feature extraction and relation modeling by jointly encoding template and search tokens, as in OSTrack~\cite{ye_2022_joint} and MixFormer~\cite{Cui_2022_MixFormer}. 
Recent trackers extend the matching paradigm with autoregressive prediction~\cite{Chen_2023_CVPR_seqtrack,Wei_2023_CVPR_autoregressive}, online query propagation~\cite{ODTrack,Xie_2024_CVPR_AQATrack}, historical prompts~\cite{Cai_2024_CVPR_HIPTrack}, and richer video-level context propagation~\cite{kang2025exploring_mcitrack,li2025mambalct}. 
However, their evidence is still concentrated in one prediction stream, so occlusion, similar distractors, or unreliable historical states can produce temporally smooth but semantically incorrect boxes. 
ACTrack therefore treats such trackers as replaceable \emph{Instance Matching Tools}, whose box-level evidence can be coordinated with complementary tools.

\subsection{Tracking with Visual Foundation Models}

Visual foundation models provide transferable representations and promptable perception capabilities that go beyond tracking-only training. One line of work adapts pretrained visual backbones into trackers. LoRAT applies low-rank adaptation to visual tracking, making it possible to train larger pretrained backbones with reduced cost~\cite{LoRAT}. SPMTrack further explores spatio-temporal parameter-efficient tuning and mixture-of-experts for scalable tracking~\cite{Cai_2025_CVPR_SPMTrack}. HIPTrack shows that high-quality historical prompts can improve tracking by injecting target history into the tracking process~\cite{Cai_2024_CVPR_HIPTrack}. These methods use foundation models mainly as strong representation backbones or promptable tracking components.

Another line uses segmentation foundation models as tracking components. SAM2 extends promptable segmentation from images to videos with a streaming memory mechanism~\cite{ravi2025sam2}, and SAM3 further introduces concept-level promptable segmentation~\cite{carion2026sam3segmentconcepts}. SAM-based trackers improve their suitability for visual tracking by adding tracking-oriented memory or motion modeling. SAMURAI introduces motion-aware memory selection for zero-shot visual tracking with SAM2~\cite{yang2026samurai}, while DAM4SAM and SAM2.1++~\cite{Videnovic_2025_CVPR_sam2.1++} incorporates distractor-aware memory to reduce drift under similar objects. These works show that segmentation foundation models can provide useful masks, motion cues, and foreground priors. ACTrack uses SAM3 in a different role: it is not converted into the entire tracker, but invoked as a \emph{Motion Tool} and a \emph{Perception Tool} that supplies mask-derived motion boxes, search-region anchors, and instance-conflict evidence for a matching-based tracker.

\subsection{Multimodal and Unified Multimodal Tracking}

Multimodal tracking introduces auxiliary modalities such as thermal, depth, event, or language to improve robustness under low illumination, occlusion, fast motion, and appearance ambiguity. RGB-T tracking is supported by benchmarks such as LasHeR~\cite{9640453_lasher}, RGB-D tracking by DepthTrack~\cite{Yan_2021_ICCV_depthtrack}, and RGB-Event tracking by VisEvent~\cite{10284004_visevent}. Methodologically, many multimodal trackers are designed for a specific RGB-X setting. ProTrack prompts RGB trackers with multimodal information~\cite{yang2022prompting_protrack}, ViPT uses visual prompts to adapt pretrained RGB trackers to RGB-D, RGB-T, and RGB-E tracking~\cite{Zhu_2023_CVPR_vipt}, and eMoE-Tracker focuses on event-guided tracking~\cite{chen2024emoetracker}. Other methods, such as SDSTrack~\cite{Hou_2024_CVPR_sdstrack} and OneTracker~\cite{Hong_2024_CVPR_onetracker}, learn stronger modality fusion or efficient adaptation for multimodal tracking.

Unified multimodal tracking sets a more demanding objective: one architecture, and preferably one shared parameter set, should support RGB, RGB-D, RGB-T, RGB-Event, and RGB-Language tracking rather than training separate trackers for each modality. 
SeqTrackV2 formulates tracking in RGB and other modalities in a unified sequence-to-sequence framework~\cite{chen2023unified_seqtrackv2}. 
UnTrack explicitly studies single-model and any-modality video object tracking~\cite{Wu_2024_CVPR_untrack}. 
SUTrack further targets simple and unified single object tracking across RGB and RGB-X settings~\cite{sutrack}, and Uni-MDTrack learns decoupled memory and dynamic states for all-modality tracking~\cite{cai2026unimdtracklearningdecoupledmemory}. 
ACTrack is complementary to these trackers. It does not require all modalities to be fused inside every tool; instead, multimodal capability is exposed through the Instance Matching Tool interface, while motion, perception, and reprompting tools are coordinated in the same way.

\subsection{VLMs in Tracking}


Large Vision-Language Models further enable target description generation, prompt refinement, and high-level comparison between ambiguous candidates. 
ChatTracker uses a VLM to generate high-quality target descriptions and iteratively refine ambiguous descriptions with tracking feedback~\cite{sun2024chattracker}. However, using a VLM as a dense frame-by-frame localizer is expensive and can introduce unstable decisions. 
Elysium~\cite{wang2024elysiumexploringobjectlevelperception} and Merlin~\cite{yu2024merlin} ubstantially underperforms existing trackers on tracking datasets such as LaSOT~\cite{fan2019lasot}. VPTrack~\cite{wang2026vptrackerglobalvisionlanguagetracking} leverages the language understanding capability of VLM and performs well in RGB-Language tracking, but it still significantly underperforms existing best trackers on TNL2K~\cite{Wang_2021_CVPR_TNL2k}.
Unlike previous methods, our ACTrack uses a VLM as \emph{Reprompt Tool} rather than a tracker. 
The tool is invoked only when persistent conflict suggests that visual tools disagree about the target identity, and its output is used to refresh the state in the visual tools.
This design keeps dense localization inside efficient visual tools while allowing semantic reasoning to enter the loop when the shared target state becomes unreliable.

\section{Method}
\label{sec:method}

\begin{figure*}[t]
\centering
\includegraphics[width=\textwidth]{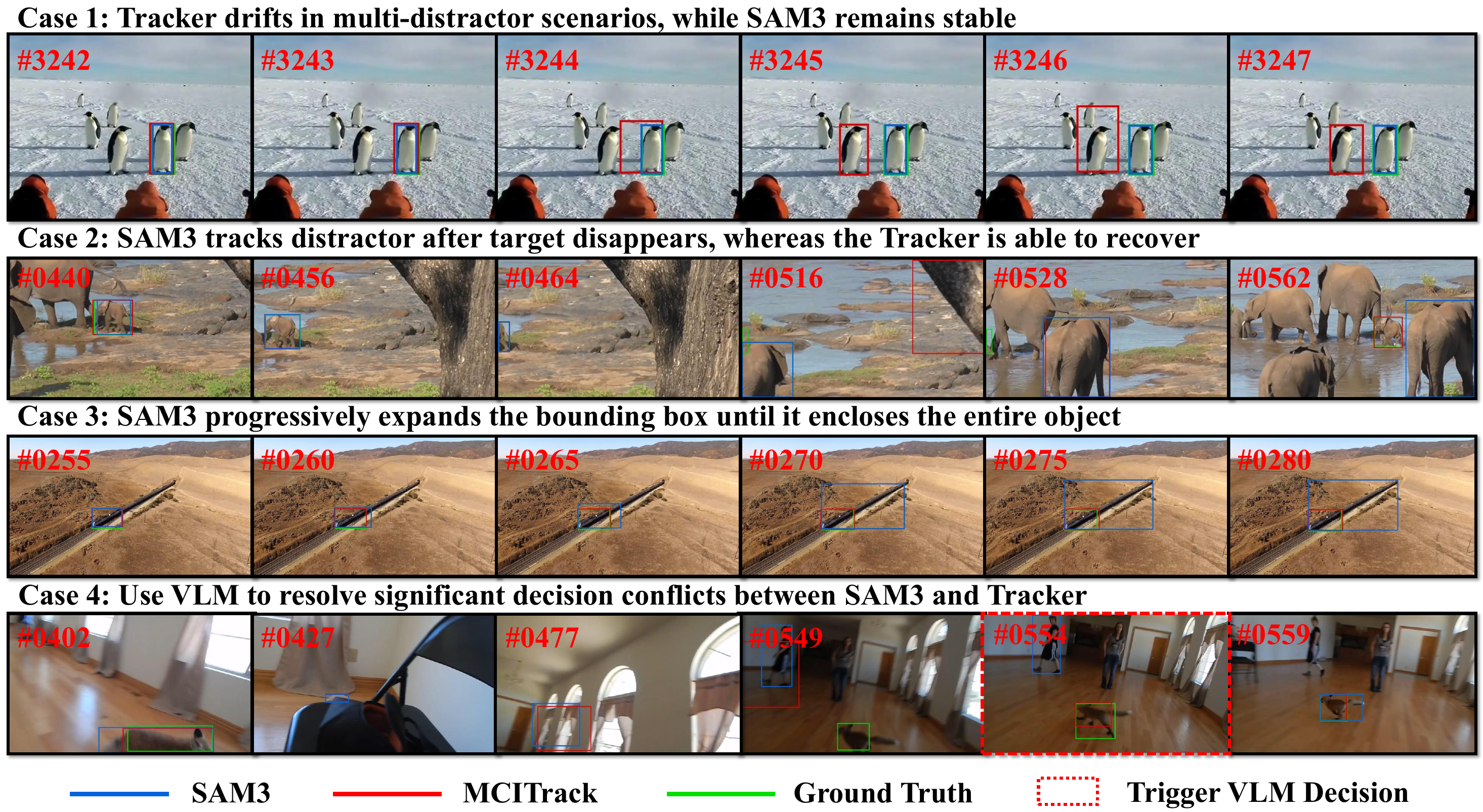}
\caption{
Case analysis of tracking failures for Tracker and SAM3 in different scenarios. 
}
\label{fig:case_analysis}
\end{figure*}

\begin{figure*}[t]
\centering
\includegraphics[width=\textwidth]{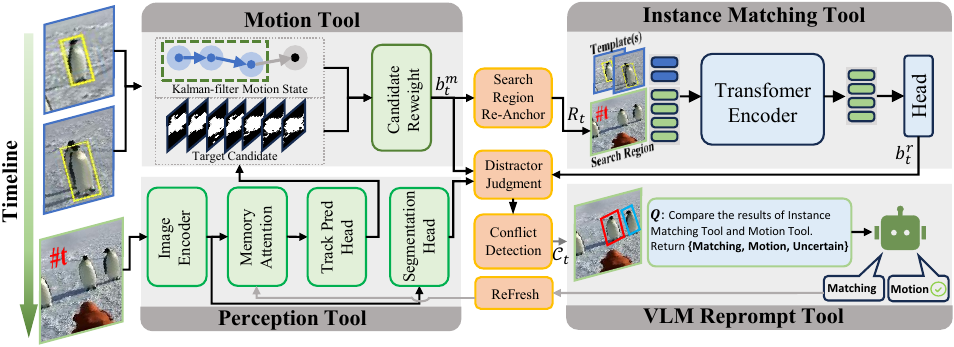}
\caption{\textbf{Overview of ACTrack as agentic tool coordination.}
ACTrack maintains a shared target state $\Omega_t$ and invokes heterogeneous visual models as tools in a closed loop. 
The SAM3 Motion Tool propagates target memory to obtain a mask-derived motion box $b_t^m$ and its search prior $R_t$. 
The Instance Matching Tool predicts the tracker-side box $b_t^r$ under this prior. 
The SAM3 Perception Tool detects similar instances in the effective search region and assigns $b_t^r$ and $b_t^m$ to instance identities, yielding the conflict signal $\mathcal{C}_t$. 
If the same conflict persists for $K$ frames, the VLM Reprompt Tool performs identity arbitration and refreshes the SAM3 prompt when $b_t^r$ is more reliable.
}
\label{fig:actrack_framework}
\end{figure*}

\subsection{Models as Tools}

This paper starts from a simple observation: 
modern visual models are not equally reliable under all tracking scenarios. 
A segmentation foundation model, a matching-based tracker, and a VLM each encode a different inductive bias. Instead of forcing all abilities into a monolithic tracker, we formulate visual tracking as \emph{agentic tool coordination}: each model is invoked for the subproblem it is good at, and its failure modes are monitored by other tools.

\textbf{SAM3 as motion and perception tools.} 
SAM-style video segmentation models provide promptable mask propagation and object-level memory, and recent SAM2/SAM3 variants further extend this ability to video and concept-conditioned segmentation~\cite{ravi2025sam2,carion2026sam3segmentconcepts}. 
As show in the first row of Fig.~\ref{fig:case_analysis}, the strength of SAM3 is foreground awareness. A propagated mask usually describes the target foreground more precisely than a bounding box, so the converted mask box provides a clean search prior for later matching.
This is especially useful when background pixels inside a tracker crop become misleading. SAM3 is also useful in multi-distractor scenes because it can return object-level masks for similar instances, making it possible to reason about whether two boxes correspond to the same physical object. 
Motion-aware and distractor-aware SAM variants also show that obtaining the foreground mask of the target is  conducive to estimating the motion state, and explicit motion and memory design is important for visual tracking with segmentation models~\cite{Videnovic_2025_CVPR_sam2.1++,yang2026samurai}.

However, a segmentation model does not fully solve single-object tracking. Firstly, as shown in the last three rows of Fig.~\ref{fig:case_analysis}, promptable segmentation follows the semantically meaningful object extent, while tracking annotations define a specific target instance that may correspond to a part, the whole, or only a subset of a semantic object. This mismatch leads to scale drift in both directions. The mask may expand from a part to the whole semantic object: if the initial box covers one carriage of a train, segmentation memory tends to grow to the entire train because the whole train is a coherent visual object, while the tracker should preserve the annotated carriage identity. The mask may also collapse from the whole object to a discriminative local region: under occlusion, motion blur, or competition from similar instances, a small high-contrast patch becomes a more stable anchor for memory attention than the full object outline. 
Secondly, when the target disappears, a segmentation propagation tool may output no reliable mask; after the target reappears, it may be unable to re-lock without a new prompt. Thirdly, once a wrong object, an over-expanded mask, or a collapsed partial mask is written into the video memory, the memory pool can be polluted and future masks may reinforce the error.

\textbf{Trackers as instance matching tools.} 
Matching-based trackers formulate tracking as matching a target template to a search region~\cite{bertinetto2016fully,li2018high,li2019siamrpn++,yan2021learning,ye_2022_joint,Cui_2022_MixFormer}. Compared with a pure mask propagator, they usually preserve the annotated instance more faithfully: the tracker learns to follow the target specified by the first-frame crop or the additional context, as shown in the third row of Fig.~\ref{fig:case_analysis}. Trackers also tend to continue exploring when the target becomes unreliable, as demonstrated in the second row of Fig.~\ref{fig:case_analysis}. This does not make it easy to deal with disappearance scenarios, but in difficult out-of-view or reappearance scenarios, a tracker can still test candidate regions instead of simply returning a distractor or no object.

The weakness of a tracker is that its crop and response map can be distracted by nearby similar objects, as shown in the first row of Fig.~\ref{fig:case_analysis}. Current methods often apply Hanning windows~\cite{bertinetto2016fully} to introduce motion prior and stabilize the response peak around the previous target center. This prior improves short-term smoothness but makes the tracker biased toward local candidates and can amplify distractor errors when a similar object stays near the search center. Stronger spatio-temporal transformer interaction reduce this issue~\cite{yan2021learning,Wei_2023_CVPR_autoregressive,Cai_2024_CVPR_HIPTrack}, but a box tracker still lacks the mask-level evidence required to tell whether two high-score boxes belong to different visible instances, as well as the precise motion state.

\textbf{VLMs as reprompt tools.} 
As mentioned in Section \ref{sec:related}, the advantage of a VLM is not low-level localization speed but semantic comparison. Given an initial reference, the current frame, and two competing hypotheses, a VLM can decide which candidate better matches the intended target identity. Another weakness of a VLM is cost and instability: dense per-frame VLM inference is unnecessary for routine localization and may introduce inconsistent semantic judgments. Therefore, our ACTrack uses the VLM only as a sparse reprompt tool under persistent disagreement. As shown in the last row of Fig.~\ref{fig:case_analysis}, when there is a serious conflict between the tracker output and the SAM3 output, the VLM intervenes in decision-making and prompts other tools to correct the track trajectory.

These complementary failure modes motivate the tool decomposition in Fig.~\ref{fig:actrack_framework}. SAM3 provides mask-level motion and perception evidence, the tracker provides instance-level matching, and the VLM provides rare semantic decision-making. ACTrack maintains a shared target state and decides which tool should influence the next frame.

\subsection{Overview: Agentic Coordination Framework}

This section details how ACTrack synchronizes and coordinates its tools into a closed-loop tracking agent, as illustrated in Fig.~\ref{fig:actrack_framework}. We process a video frame by frame: at step $t$ the input is an RGB frame $I_t\in\mathbb{R}^{H\times W\times 3}$ (optionally with an auxiliary modality, Section~\ref{sec:peft_mm}), and the goal is to predict the target box $\bar{b}_t=(x,y,w,h)\in\mathbb{R}^{4}$ specified by the first-frame annotation $(I_0,b_0)$.
We write $\operatorname{Expand}(b,\rho)$ for a box scaled around its center by a factor $\rho$.

Rather than fusing tool outputs by a fixed rule, ACTrack maintains a shared target state
\begin{equation}
\Omega_t=\{\bar{b}_{t-1},\, H_{t-1},\, S_{t-1},\, Q_{t-1},\, (I_0,b_0)\},
\end{equation}
where $\bar{b}_{t-1}\in\mathbb{R}^{4}$ is the last accepted box, 
$H_{t-1}$ denotes the implementation-dependent state of the Instance Matching Tool, including its target state, template or memory features, and any hidden states or caches used by the underlying tracker (Section~\ref{sec:matching}), 
$S_{t-1}$ denotes the SAM3 video state, including mask memories and object-pointer tokens used for propagation (Section~\ref{sec:motion_perception}),
$Q_{t-1}\in\{0,1\}^{K}$ denotes a sliding buffer of recent conflict flags, 
and $(I_0,b_0)$ is the immutable initial reference. 
At each frame, ACTrack invokes the tools in a fixed order: Motion, Instance Matching, and Perception, with Reprompt called only when its trigger is met. Each tool reads from and writes back to $\Omega_t$, turning three heterogeneous models into one feedback loop.

\begin{figure*}[!t]
\centering
\includegraphics[width=\textwidth]{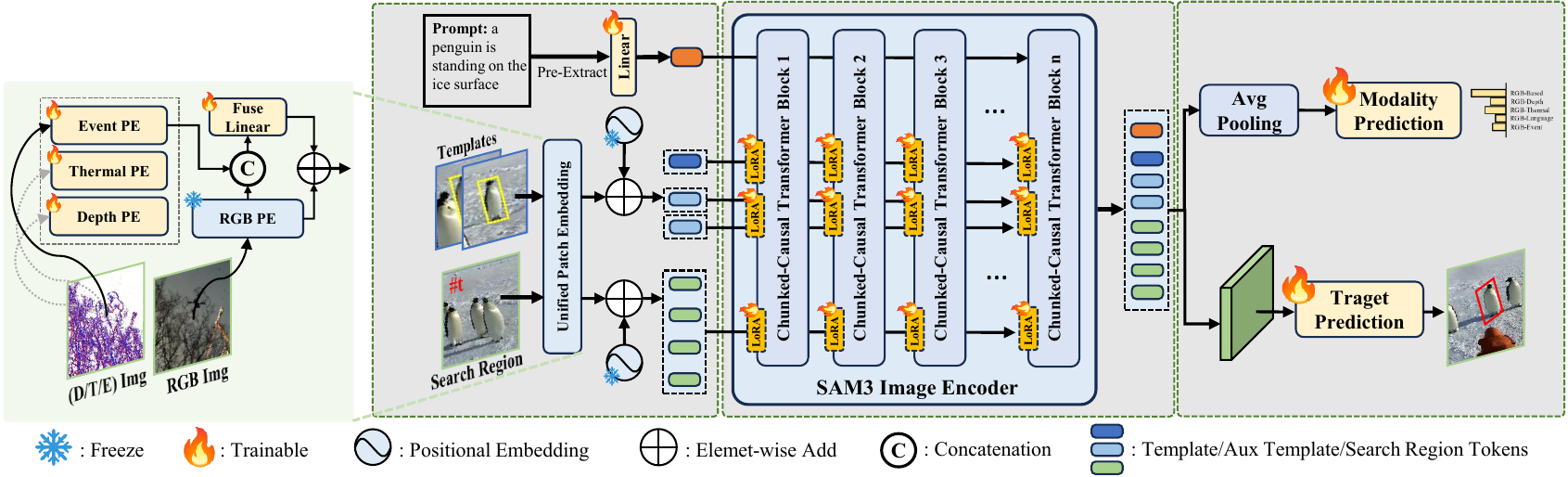}
\caption{Architecture of the parameter-efficient unified multimodal Instance Matching Tool in ACTrack-EM. RGB and auxiliary inputs are first encoded by modality-aware patch embeddings with residual fusion. The resulting template, auxiliary-template, and search-region tokens are processed by the frozen SAM3 image encoder equipped with stream-specific LoRA paths. The tracker predicts the target box from the search tokens, while an auxiliary modality head regularizes the unified representation across RGB, RGB-D, RGB-T, and RGB-Event inputs.}
\label{fig:actrack_em}
\end{figure*}

\textbf{Motion Tool provides search region priors for the Instance Matching Tool.}
The SAM3 Motion Tool propagates a mask to frame $I_t$ and converts it to a motion box $b_t^m\in\mathbb{R}^{4}$.
This box is used as a motion prior to \emph{re-anchor} the search center of the Instance Matching Tool, and is expanded by $R_t=\operatorname{Expand}(b_t^m,\rho)$ to construct the effective search region.
Therefore, the matched box $b_t^r\in\mathbb{R}^{4}$ predicted by the Instance Matching Tool are steered by mask-level motion evidence while keeping the tracker's own discriminative search. 
If no valid mask is returned, ACTrack falls back to using $\bar{b}_{t-1}$ to crop the search region and skips the perception stage.

\textbf{Perception Tool provides judgments of distractors within the search region.}
Given $R_t$, the SAM3 Perception Tool detects instances inside the search region and assigns both $b_t^r$ and $b_t^m$ to the detected instances, yielding identities $a_t^r$ and $a_t^m$. 
A reliable conflict $\mathcal{C}_t=\mathbb{1}[a_t^r\neq a_t^m]\in\{0,1\}$ signals that the matcher and the motion prior have locked onto \emph{different physical objects}; 
this instance-level evidence, not a scalar IoU, drives all subsequent coordination.

\textbf{Default Decision-Making Policy.}
When conflict-free ($\mathcal{C}_t=0$), ACTrack accepts the box output by the Instance Matching Tool $\bar{b}_t=b_t^r$, since the tracker preserves the annotated instance most faithfully; under a reliable conflict it temporarily favors the box $b_t^m$ output by the Motion Tool to avoid contaminating the tracker with a distractor. The accepted $\bar{b}_t$ is committed back to both $H_t$ and $S_t$, so each decision conditions the next frame.

\textbf{Reprompt Tool is triggered by persistent disagreement.}
A one-frame conflict is treated as transient evidence. ACTrack queries the VLM Reprompt Tool only when the conflict buffer $Q_t$ records the same conflict for $K$ consecutive frames, obtaining a verdict $y_t\in\{\textsc{Matching},\textsc{Motion},\textsc{Uncertain}\}$.
If $y_t=\textsc{Matching}$, the VLM judges that the Instance Matching Tool still follows the annotated target while SAM3 has drifted. ACTrack therefore accepts $b_t^r$, injects it into SAM3 as a new box prompt at the current frame, and restarts SAM3 mask propagation from the next frame. If $y_t=\textsc{Motion}$, ACTrack keeps the motion path $b_t^m$ and does not overwrite the SAM3 state with the matcher output. If $y_t=\textsc{Uncertain}$, ACTrack leaves the default decision unchanged and avoids forcing either tool to overwrite the other. The VLM is therefore used only for sparse identity arbitration and prompt renewal, not for dense localization.
\subsection{Motion Tool and Perception Tool}
\label{sec:motion_perception}

ACTrack uses SAM3 in two roles. As the \emph{Motion Tool}, SAM3 propagates the target memory in $S_{t-1}$ to frame $I_t$, and its mask decoder yields $N$ candidate masks $\{M_t^{(i)}\}$ with mask-quality scores $u_t\in\mathbb{R}^{N}$ (the IoU-prediction head) and an objectness logit $o_t$. The selected mask is binarized into the motion box $b_t^m\in\mathbb{R}^{4}$ and expanded into $R_t=\operatorname{Expand}(b_t^m,\rho)$, where $\rho$ is the expansion rate. If no valid mask is returned, ACTrack falls back to $\bar{b}_{t-1}$ and skips conflict detection. The concrete hyperparameter values are given in Section~\ref{sec:experiments}.

\textbf{Kalman-filtered mask selection.}
A purely appearance-driven decoder ranks candidates only by $u_t$ and tends to switch identity in clutter. Following motion-aware SAM trackers~\cite{yang2026samurai}, we inject a motion prior, but re-rank the \emph{mask candidates} before they enter the SAM3 memory rather than re-ranking final boxes, which keeps a distractor mask from polluting the video state. We maintain a constant-velocity Kalman filter (KF) whose state $\mathbf{x}_t\in\mathbb{R}^{8}$ stacks the box center, aspect ratio, height, and their velocities, with measurement $(c_x,c_y,w/h,h)$ read from the selected mask box and noise scaled by the target height~\cite{wojke2017simple}. Converting each candidate mask to a box and computing its IoU $g_t^{(i)}$ with the KF-predicted box $\hat{b}_t$, we select the mask by a motion-aware score
\begin{equation}
i^{\star}=\arg\max_{i}\;\big[(1-W_\mathrm{m})\,u_t^{(i)} + W_\mathrm{m}\,g_t^{(i)}\big],
\label{eq:kf_fusion}
\end{equation}
where $W_\mathrm{m}$ controls the contribution of the motion prior. The motion term is enabled only after a warm-up period with consistently reliable mask predictions; otherwise the selection falls back to $u_t$ alone. The objectness logit gates the whole process: when the target is deemed absent, the masks are suppressed, and the KF state is cleared so a stale trajectory is not carried through an occlusion. Eq.~\eqref{eq:kf_fusion} only changes which mask is kept; it adds no learnable parameter and leaves SAM3 frozen.

\textbf{Perception Tool.}
As the \emph{Perception Tool}, SAM3 uses its image detection and segmentation branch to detect similar objects inside the effective search region, 
providing instance-level evidence rather than a better IoU. 
Let $\mathcal{O}_t=\{(B_i,M_i,q_i)\}_{i=1}^{N_t}$ be the detected instances, 
where $B_i\in\mathbb{R}^{4}$ is the bounding box of instance $i$ in the original image coordinates, 
$M_i\in\{0,1\}^{H_R\times W_R}$ is its binary mask in the effective-search-region crop, 
and $q_i$ is the detection confidence. 
ACTrack then applies the assignment rule in Algorithm~\ref{alg:perception_assignment} to both the Instance Matching box $b_t^r$ and the Motion Tool box $b_t^m$. 
The assignment uses the centers of both boxes to query the SAM3 detections. When a valid instance mask is available, ACTrack checks whether the box center lies inside that mask; otherwise it falls back to containment by the detected box $B_i$. 
We denote by $\Pi_R(\cdot)$ the coordinate mapping from an original-image point to the mask grid of the crop specified by search region $R$: it subtracts the crop's top-left corner and scales the point to the $H_R\times W_R$ mask resolution. For a candidate set $\mathcal{I}$ and a query center $c$, $\operatorname{SelectBest}(\mathcal{I},c)$ first selects the instance with the highest confidence $q_i$; if multiple instances have the same confidence, it selects the one with the smallest center distance $\|c-\operatorname{Center}(B_i)\|_2$.
A conflict means the two tools have locked onto different physical objects. This exposes the mismatch between semantic object perception and tracking identity, and ACTrack must decide how to update the shared state.

\begin{algorithm}[t]
\small
\caption{Perception-based instance assignment}
\label{alg:perception_assignment}
\begin{algorithmic}[1]
\Require Search region $R_t$, boxes $b_t^r,b_t^m$
\Require SAM3 detections $\mathcal{O}_t=\{(B_i,M_i,q_i)\}_{i=1}^{N_t}$
\Require $B_i$: image-coordinate box;\quad $M_i$: mask on the $R_t$ crop grid
\Ensure Assignments $a_t^r,a_t^m$ and conflict flag $\mathcal{C}_t$
\Function{Assign}{$b,\mathcal{O}_t,R_t$}
    \State $c\gets \operatorname{Center}(b)$
    \State $\tilde{c}\gets \Pi_{R_t}(c)$
    \State $\mathcal{I}_m\gets\{i\mid \tilde{c}\in M_i\}$
    \If{$\mathcal{I}_m\neq\emptyset$}
        \State \Return $\operatorname{SelectBest}(\mathcal{I}_m,c)$
    \EndIf
    \State $\mathcal{I}_b\gets\{i\mid c\in B_i\}$
    \If{$\mathcal{I}_b\neq\emptyset$}
        \State \Return $\operatorname{SelectBest}(\mathcal{I}_b,c)$
    \EndIf
    \State \Return $\varnothing$
\EndFunction
\State $a_t^r\gets \operatorname{Assign}(b_t^r,\mathcal{O}_t,R_t)$
\State $a_t^m\gets \operatorname{Assign}(b_t^m,\mathcal{O}_t,R_t)$
\State $v\gets(a_t^r\neq\varnothing)\land(a_t^m\neq\varnothing)$
\State $\mathcal{C}_t\gets \mathbb{1}[v\land a_t^r\neq a_t^m]$
\State \Return $a_t^r,a_t^m,\mathcal{C}_t$
\end{algorithmic}
\end{algorithm}

\subsection{Instance Matching Tool}
\label{sec:matching}

The Instance Matching Tool is the default localization engine. Given the current frame $I_t$, the tracker state $H_{t-1}$, and the search prior $R_t$ from the Motion Tool, it predicts the tracker-side target hypothesis and updates its internal state:
\begin{equation}
(b_t^r, H_t^r) = \mathcal{D}(I_t, R_t, H_{t-1}),
\end{equation}
where $b_t^r\in\mathbb{R}^{4}$ is the box predicted by the Instance Matching Tool, and $H_t^r$ is the tracker state after matching. The interface requires three operations: initialization with the first-frame target, frame-wise matching under the motion-guided search prior, and state synchronization after ACTrack selects the accepted box. The coordination policy does not rely on a scalar tracker confidence; instead, the tracker output is checked against SAM3's instance-level evidence by the Perception Tool. The role of $R_t$ is to constrain the search process with mask-level motion evidence, while the tracker still performs its own discriminative matching using its templates and temporal state.

Because the tool is defined only by this box-and-state interface, any compatible tracker can be plugged into ACTrack; for RGB modality, we use an off-the-shelf matching-based tracker, and Section~\ref{sec:peft_mm} provides a parameter-efficient tuned unified multimodal tracker $\mathcal{D}_{mm}$. 
Our contribution is therefore not a new tracker followed matching paradigm but a coordination policy that lets a matching-based tracker interact with segmentation, perception, and semantic reasoning tools.

\subsection{VLM Reprompt Tool}
\label{sec:vlm}

A single conflict flag $\mathcal{C}_t$ is unreliable, so ACTrack escalates to semantic reasoning only when the disagreement is \emph{persistent}: the VLM is queried only when the conflict buffer $Q_t$ holds the same conflict for $K$ consecutive frames. 
This keeps the average number of VLM calls per sequence small, so the semantic cost stays negligible relative to per-frame localization (Section~\ref{sec:experiments}).

When triggered, the VLM receives a four-panel image. The panel contains the initial reference from $(I_0,b_0)$, the current frame with both $b_t^r$ and $b_t^m$ overlaid, and one crop around each box. The VLM then returns a three-way verdict
\begin{equation}
\begin{aligned}
y_t &= \mathcal{V}\big((I_0,b_0),\,I_t,\,b_t^r,\,b_t^m\big), \\
y_t &\in \{\textsc{Matching},\,\textsc{Motion},\,\textsc{Uncertain}\}.
\end{aligned}
\end{equation}
\textsc{Matching} means the Instance Matching Tool still follows the annotated target while SAM3 has drifted. ACTrack accepts $b_t^r$, injects it into SAM3 as a new box prompt at the current frame, and restarts mask propagation from the next frame. \textsc{Motion} keeps the motion path $b_t^m$, and \textsc{Uncertain} leaves the default decision unchanged. The VLM is used only for sparse identity arbitration and prompt renewal, not for dense localization.

\subsection{The Coordination Algorithm of ACTrack}
\label{sec:coordination}

Fig.~\ref{fig:actrack_framework} illustrates the closed-loop interaction among the tools, and Algorithm~\ref{alg:actrack} gives the corresponding per-frame control flow.
The Motion Tool first provides a mask-derived motion prior, and the Instance Matching Tool uses this prior to produce the default target hypothesis. The Perception Tool then checks whether the Instance Matching box and the motion box belong to the same detected instance. 
On ordinary frames, ACTrack follows the default policy directly: it accepts $b_t^r$ when there is no conflict and temporarily favors $b_t^m$ under a reliable conflict. 
Only when the same conflict persists in $Q_t$ for $K$ frames does ACTrack query the VLM Reprompt Tool. 
In the algorithm, $\eta_t$ indicates whether it returns a valid mask, $\operatorname{PerceptionAssign}$ refers to Algorithm~\ref{alg:perception_assignment}, and $\operatorname{Persistent}(Q_t)$ denotes the repeated-conflict trigger. The final accepted box is committed back to both tool states, so the next frame is conditioned on the current decision.

\begin{algorithm}[t]
\small
\caption{ACTrack Inference}
\label{alg:actrack}
\begin{algorithmic}[1]
\Require Frame $I_t$, accepted box $\bar{b}_{t-1}$
\Require Tool states $H_{t-1},S_{t-1}$, buffer $Q_{t-1}$, reference $(I_0,b_0)$
\Ensure Accepted box $\bar{b}_t$, updated state $\Omega_{t+1}$
\State $(M_t,b_t^m,\eta_t)\gets \operatorname{MotionTool}(I_t,S_{t-1})$
\If{$\eta_t=0$}
    \State $b_t^m\gets \bar{b}_{t-1}$
    \State $\mathcal{C}_t\gets0$
\EndIf
\State $R_t\gets \operatorname{Expand}(b_t^m,\rho)$
\State $(b_t^r,H_t^r)\gets \mathcal{D}(I_t,R_t,H_{t-1})$
\If{$\eta_t=1$}
    \State $\mathcal{O}_t\gets \operatorname{PerceptionTool}(I_t,R_t)$
    \State $(a_t^r,a_t^m,\mathcal{C}_t)\gets$
    \Statex \hspace{\algorithmicindent}$\operatorname{PerceptionAssign}(R_t,b_t^r,b_t^m,\mathcal{O}_t)$
\EndIf
\State $Q_t\gets \operatorname{Push}(Q_{t-1},\mathcal{C}_t)$
\If{$\mathcal{C}_t=0$}
    \State $\bar{b}_t\gets b_t^r$
\Else
    \State $\bar{b}_t\gets b_t^m$
\EndIf
\State $S_t\gets S_{t-1}$
\If{$\operatorname{Persistent}(Q_t)$}
    \State $y_t\gets \mathcal{V}((I_0,b_0),I_t,b_t^r,b_t^m)$
    \If{$y_t=\textsc{Matching}$}
        \State $\bar{b}_t\gets b_t^r$
        \State $S_t\gets \operatorname{Reprompt}(S_{t-1},b_t^r)$
    \ElsIf{$y_t=\textsc{Motion}$}
        \State $\bar{b}_t\gets b_t^m$
    \EndIf
\EndIf
\State $(H_t,S_t)\gets \operatorname{Commit}(H_t^r,S_t,\bar{b}_t)$
\State \Return $\bar{b}_t,\ \Omega_{t+1}$
\end{algorithmic}
\end{algorithm}

\subsection{Unified Multimodal Instance Matching Tool with Parameter-Efficient Adaptation}
\label{sec:peft_mm}

The ACTrack coordination policy is defined over tool outputs, not over a particular image modality. 
This makes multimodal tracking a natural extension of the framework: the Motion, Perception, and Reprompt Tools keep their original interfaces, while modality-specific fusion is isolated inside the Instance Matching Tool. 
ACTrack-EM implements this idea by parameter-\underline{e}fficiently adapting the SAM3 image encoder into a unified \underline{m}ultimodal Instance Matching Tool that absorbs heterogeneous RGB-X inputs and still returns the same box-and-state output to ACTrack. 
The SAM3 backbone $\Phi_0$ is kept frozen, and only lightweight modality-specific patch embedding, stream-specific parameters, and prediction head are learnable. 
This design reuses the visual foundation backbone across tools, avoids training separate trackers for different modalities, and keeps the additional parameter cost small.

Each template, auxiliary-template, or search region is cropped and resized before patch embedding. We write the resulting input as $X=[X^{rgb};X^{aux}]\in\mathbb{R}^{6\times H\times W}$; with patch size $p$, it yields $n_{H,W}=(H/p)(W/p)$ visual tokens. Here $X^{aux}$ denotes the auxiliary modality, such as depth, thermal, or event data, and is replaced by a copy of RGB when no auxiliary modality is available. A task index $\kappa$ indicates the input modality and also supervises an auxiliary modality-prediction head during training. As illustrated in Fig.~\ref{fig:actrack_em}, the adaptation consists of modality-aware unified patch embedding, dynamic auxiliary templates, stream-specific low-rank updates, and causal temporal token interaction.

\textbf{Modality-aware unified patch embedding.}
Instead of feeding all modalities through a single projection, ACTrack-EM embeds the RGB channels with the frozen patch embedding $\mathrm{PE}_{rgb}$ and the auxiliary channels with a modality-specific expert $\mathrm{PE}_{\kappa}$ initialized from $\mathrm{PE}_{rgb}$. This gives $E_{rgb},E_{aux}\in\mathbb{R}^{d\times (H/p)\times (W/p)}$. The two streams are merged as $E=E_{rgb}+\operatorname{Fuse}([E_{rgb};E_{aux}])$, so training starts from the pretrained RGB representation and the auxiliary cue enters as a learned residual. For RGB-only inputs, the auxiliary branch receives the RGB copy, allowing the same module to serve both RGB and RGB-X settings. Flattening $E$ gives the token sequence in $\mathbb{R}^{n_{H,W}\times d}$.

\textbf{Dynamic auxiliary templates.}
Besides the fixed initial template $Z_0$ cropped from $(I_0,b_0)$, ACTrack-EM maintains a FIFO queue of dynamic auxiliary templates to describe appearance changes over time. These templates are encoded as additional causal chunks between the initial template and the search region. At inference, the underlying tracker uses its predicted target confidence to decide when the auxiliary-template queue should be refreshed. This confidence is used only for template management inside the Instance Matching Tool and is not part of ACTrack's cross-tool coordination policy.

\textbf{Multi-path stream-specific LoRA.}
The backbone processes token streams with different roles: the initial-template stream, the dynamic-template stream, and the search-region stream. We therefore avoid sharing one adaptation path across all streams. Each linear projection $W$ in attention and MLP is equipped with stream-specific LoRA experts, written as $W'_j = W + \frac{\alpha}{r}B_jA_j$. Here $j$ indexes one of the streams above; all dynamic auxiliary templates share the dynamic-template expert, while the search-region tokens are routed to the search expert. This static routing lets the search stream specialize for localization without changing the template representation learned from the reference frames, while the SAM3 backbone weights remain frozen~\cite{lin2025lorat_v2}.

\textbf{Chunked causal temporal attention.}
The original image encoder processes a single image crop, whereas tracking requires interaction among the initial template, auxiliary templates, and search frame. ACTrack-EM orders these token groups as temporal chunks and uses chunked causal attention to fuse them. When a language description is available, we use its pre-extracted T5 embedding and project it with a lightweight MLP into an NLP prefix token placed before the initial-template tokens; otherwise a default placeholder token is used. This prefix token is context input rather than a separate LoRA stream. Tokens in a later chunk can attend to previous chunks and to their own chunk, while earlier chunks are not updated by future search tokens. During inference, the NLP prefix and template chunks are encoded into a KV cache, and the search chunk reads from the cached template context. This preserves the temporal direction of tracking and avoids contaminating the reference representation with information from later frames.

Together, $\mathcal{D}_{mm}$ exposes exactly the same box-and-state interface $(b_t^r,H_t^r)=\mathcal{D}_{mm}(I_t,R_t,H_{t-1})$. As a result, ACTrack can use one coordination policy for RGB and RGB-X tracking: modality-specific evidence is handled inside the Instance Matching Tool, while motion propagation, instance-level perception, and sparse VLM reprompting remain unchanged.

\section{Experiments}
\label{sec:experiments}
\newcommand{\first}[1]{\textbf{\textcolor{red}{#1}}}
\newcommand{\second}[1]{\textbf{\textcolor{blue}{#1}}}
\newcommand{\third}[1]{\textbf{#1}}

\subsection{Implementation Details}

\noindent\textbf{Model settings.}
In our experiments, we evaluated ACTrack under three configurations. All configurations maintain the same agentic coordination framework, with the only difference being the Instance Matching tool used. For the Motion Tool, we adopt a SAM3 implementation enhanced with Kalman filter; for the Perception Tool, we use the original SAM3 implementation; and for the Reprompt Tool, we default to Seed 2.0 Pro~\cite{seed2026seed2}.
For classic RGB tracking, we adopt two configurations: ACTrack-B$_{224}$ built upon MCITrack-B$_{224}$, and ACTrack-L$_{384}$ built upon MCITrack-L$_{384}$.
For the two configurations, the Instance Matching Tool uses templates of size $112\times 112$ and $192\times 192$ respectively, with corresponding search regions of $224\times 224$ and $384\times 384$. The cropping factors of the template is 2.0, and the search region is cropped based on the output of the Motion Tool. 
The inputs to the Motion Tool is the native video frames, whereas the Perception Tool applies the SAM3 image branch to the effective search-region crop for instance-conflict detection.
For unified multimodal tracking, our variant is ACTrack-EM, which shares identical components with the RGB-based ACTrack except for the Instance Matching Tool. We apply parameter-efficient fine-tuning to the image encoder of SAM3 to train the Instance Matching Tool, which consists of 31 transformer layers with a hidden size of 1024 and an MLP inner dimension of 4736. Specifically, we employ LoRA with a rank of $r=64$ and a scaling factor of $\alpha = 64$ to fine-tune the backbone. ACTrack-EM uses a $224\times 224$ initial template and two $224\times 224$ dynamic templates, along with a $378\times 378$ search region.
Table \ref{tab:model_settings} details the number of parameters, computational cost, and average inference speed of ACTrack-B$_{224}$, ACTrack-L$_{384}$, and ACTrack-EM, with all evaluations conducted on a single NVIDIA RTX 4090 GPU. The speed test is conducted on the entire LaSOT \emph{test} split and yielded an average value. The results demonstrate that integrating multiple specialized models as tools within a unified agentic coordination framework yields a significant advantage in terms of parameter count compared to previous large or giant scale trackers. Furthermore, benefiting from the parameter-efficient fine-tuning strategy, ACTrack-EM exhibits an even more pronounced advantage in parameter efficiency. In terms of inference speed, ACTrack shows no clear disadvantage against previous large-scale trackers and is substantially faster than SPMTrack-G.

\begin{table}[h]\footnotesize
    \centering
    \caption{ Comparison of our method with other trackers using parameter-efficient training method in terms of total parameters, trainable parameters, computational complexity, and inference speed. Speed is measured on an NVIDIA RTX 4090 for ACTrack and SPMTrack-G; the speed of LoRAT-G$_{378}$ is copied from the original paper.}
    \setlength{\tabcolsep}{1.5pt}
    \renewcommand{\arraystretch}{1.05}
    \begin{tabular*}{\columnwidth}{@{\extracolsep{\fill}}lcccc}
    \toprule
        \shortstack{\textbf{Method}\\[-1pt]~} & \shortstack{\textbf{Trainable}\\\textbf{Params(M)}} & \shortstack{\textbf{Params(M)}\\[-1pt]~} & \shortstack{\textbf{FLOPs(G)}\\[-1pt]~} & \shortstack{\textbf{Speed}\\\textbf{(FPS)}} \\
        \midrule
        \textbf{ACTrack-B$_{224}$} & 0 & 613.5 & 2348.09 & 15.1 \\
        \textbf{ACTrack-L$_{384}$} & 0 & 865.0 & 2679.93 & 12.7 \\
        \textbf{ACTrack-EM} &  173.3 & 559.0 & 2876.54 & 14.2 \\
        \midrule
        \textbf{SPMTrack-G} & 204.0 & 1339.5 & 3942.2 & 8.6 \\
        \textbf{LoRAT-G$_{378}$} & 80.2 & 1215.7 & 1161.0 & 20.0 \\
    \bottomrule
    \end{tabular*}
    \label{tab:model_settings}
    \vspace{-2pt}
\end{table}

\begin{table*}[t]
\centering
\caption{State-of-the-art comparison on RGB tracking benchmarks. Public baseline numbers are taken from published comparison tables. SAM-based and matching-based trackers are separated by type, while ACTrack is left ungrouped as an agentic coordination method. The best three results for each metric are highlighted in \textbf{\textcolor{red}{red}}, \textbf{\textcolor{blue}{blue}}, and \textbf{bold}, respectively.}
\label{tab:rgb_lasot_trackingnet}
\footnotesize
\resizebox{\textwidth}{!}{
\begin{tabular}{l|c|ccc|ccc|ccc}
\toprule
\textbf{Method} & \textbf{Source} & \multicolumn{3}{c|}{\textbf{LaSOT}} & \multicolumn{3}{c|}{\textbf{TrackingNet}} & \multicolumn{3}{c}{\textbf{LaSOT$_\mathrm{ext}$}} \\
\cmidrule(lr){3-5}\cmidrule(lr){6-8}\cmidrule(lr){9-11}
 & & \textbf{AUC} & $\mathbf{P_\mathrm{Norm}}$ & $\mathbf{P}$ & \textbf{AUC} & $\mathbf{P_\mathrm{Norm}}$ & $\mathbf{P}$ & \textbf{AUC} & $\mathbf{P_\mathrm{Norm}}$ & $\mathbf{P}$ \\
\midrule
\textbf{ACTrack-B$_{224}$} & Ours & \second{80.1} & \second{90.3} & \second{87.8} & 86.8 & 91.5 & 87.0 & \third{65.4} & \second{77.5} & \third{73.1} \\
\textbf{ACTrack-L$_{384}$} & Ours & \first{81.4} & \first{90.8} & \first{89.4} & \first{88.0} & \first{92.4} & \first{89.5} & \first{66.5} & \first{77.9} & \second{73.6} \\
\textbf{ACTrack-EM} & Ours & \third{79.3} & \third{88.9} & \third{87.2} & \third{87.5} & \third{92.1} & \third{88.4} & \second{65.8} & \third{75.5} & \first{76.8} \\
\midrule
\multicolumn{11}{l}{\emph{SAM-based trackers}} \\
SAMITE-B~\cite{xu2025samite} & arXiv25 & 74.9 & 83.4 & 81.4 & 84.5 & -- & -- & 60.7 & 73.1 & 71.2 \\
SAMURAI-L~\cite{yang2026samurai} & TIP26 & 74.2 & 82.7 & 80.2 & 85.3 & -- & -- & 61.0 & 73.9 & 72.2 \\
SAM2.1-L~\cite{ravi2025sam2} & ICLR25 & 68.5 & 76.2 & 73.6 & -- & -- & -- & 58.6 & 71.1 & 68.8 \\
\midrule
\multicolumn{11}{l}{\emph{Matching-based trackers}} \\
RELO-L$_{256}$~\cite{chen2026reloreinforcementlearninglocalize} & ICML26 & 75.1 & 85.1 & 83.4 & 87.3 & 91.6 & 88.0 & 57.5 & 69.1 & 66.7 \\
SPMTrack-L~\cite{Cai_2025_CVPR_SPMTrack} & CVPR25 & 76.8 & 85.9 & 84.0 & 86.9 & 91.0 & 87.2 & -- & -- & -- \\
MCITrack-L$_{384}$~\cite{kang2025exploring_mcitrack} & AAAI25 & 76.6 & 86.1 & 85.0 & \second{87.9} & \second{92.1} & \second{89.2} & 55.7 & 66.5 & 62.9 \\
LoRAT-g$_{378}$~\cite{LoRAT} & ECCV24 & 76.2 & 85.3 & 83.5 & 86.0 & 90.2 & 86.1 & 56.5 & 69.0 & 64.9 \\
LoRAT-L$_{378}$~\cite{LoRAT} & ECCV24 & 75.1 & 84.1 & 82.0 & 85.6 & 89.7 & 85.4 & 56.6 & 69.0 & 65.1 \\
LoRATv2-L$_{378}$~\cite{lin2025lorat_v2} & NeurIPS25 & 76.1 & 85.1 & 83.1 & -- & -- & -- & -- & -- & -- \\
ARPTrack-L$_{384}$~\cite{Liang_2025_CVPR_arptrack} & CVPR25 & 74.2 & 83.4 & 81.7 & 86.6 & 91.1 & 87.4 & 54.2 & 64.4 & 61.2 \\
ODTrack-L$_{384}$~\cite{ODTrack} & AAAI24 & 74.0 & 84.2 & 82.3 & 86.1 & 91.0 & 86.7 & 53.9 & 65.4 & 61.7 \\
MambaLCT$_{384}$~\cite{li2025mambalct} & AAAI25 & 73.6 & 84.1 & 81.6 & 85.2 & 89.8 & 85.2 & 53.3 & 64.8 & 61.4 \\
ARTrackV2-L$_{384}$~\cite{Bai_2024_CVPR_ARTrackV2} & CVPR24 & 73.6 & 82.8 & 81.1 & 86.1 & 90.4 & 86.2 & 53.4 & 63.7 & 60.2 \\
ARTrack-L$_{384}$~\cite{Wei_2023_CVPR_autoregressive} & CVPR23 & 73.1 & 82.2 & 80.3 & 85.6 & 89.6 & 86.0 & 52.8 & 62.9 & 59.7 \\
SeqTrack-L$_{384}$~\cite{Chen_2023_CVPR_seqtrack} & CVPR23 & 72.5 & 81.5 & 79.3 & 85.5 & 89.8 & 85.8 & 50.7 & 61.6 & 57.5 \\
\bottomrule
\end{tabular}
}
\end{table*}

\noindent\textbf{Tool coordination configuration.}
When the Motion Tool supplies the search-region crop for the Instance Matching Tool, we adopt a crop expansion ratio of $\rho=2.5$ for RGB and RGB-Language datasets.
For RGB-D/T/E datasets, the auxiliary modality conveys more salient foreground cues than the RGB modality; therefore, we adopt a smaller cropping factor $\rho=1.5$.
Inside the Motion Tool, the multimask candidates produced by the SAM3 mask head are re-ranked through a linear fusion of the predicted-IoU score of the mask decoder and the geometric IoU between each candidate box and a Kalman filter prediction, with a motion-prior weight of $W_\mathrm{m}=0.15$. 
When the sigmoid of the objectness score $P_\mathrm{obj}$ predicted by SAM3 falls below 0.5, the target is regarded as lost and all Kalman-filter states associated with it are cleared.
Only after $15$ consecutive frames in which the SAM3 IoU prediction head outputs a score above $0.3$ does the Kalman filter complete its warm-up and resume contributing to the re-ranking.
The Perception Tool runs with the default SAM3 image-branch configuration, where both the per-query confidence threshold and the per-pixel mask threshold are set to $0.5$, and the number of object queries is $200$. If the Motion Tool and the Instance Matching Tool are assigned to different object instances for $5$ consecutive frames, the VLM Reprompt Tool is invoked to arbitrate among \textsc{Matching}, \textsc{Motion}, and \textsc{Uncertain}; only when the decision is explicitly \textsc{Matching} does ACTrack inject the current Instance Matching Tool prediction into SAM3 as a new box prompt and restart mask propagation from the next frame.


\noindent\textbf{Datasets.}
ACTrack-B$_{224}$ and ACTrack-L$_{384}$ are assembled entirely from off-the-shelf checkpoints and require no additional training. 
ACTrack-EM is trained on the training splits of LaSOT~\cite{fan2019lasot}, GOT-10k~\cite{huang2019got}, COCO~\cite{lin2014microsoft}, TrackingNet~\cite{muller2018trackingnet}, VastTrack~\cite{NEURIPS2024_ec17a52e_vasttrack}, TNL2K~\cite{Wang_2021_CVPR_TNL2k}, DepthTrack~\cite{Yan_2021_ICCV_depthtrack}, VisEvent~\cite{10284004_visevent}, and LasHeR~\cite{9640453_lasher}. 
The nine datasets are sampled with equal probability. Each sample consists of four frames, consists of an initial template, two dynamic templates, and one search frame. The frames are drawn from the same sequence with pairwise gaps uniformly sampled from $[1, 200]$. For the image-only COCO dataset, each image is replicated to form a pseudo-sequence.


\noindent\textbf{Training and Optimization.}
Our method is implemented in PyTorch 2.5.1 and trained on 8 NVIDIA H800 GPUs. 
ACTrack-EM is trained in two stages. In each stage we adopt a per-GPU batch size of $16$ and train for $50$ epochs, with $131{,}072$ samples drawn per epoch. 
In the first stage, only one $224\times 224$ initial template and one $224\times 224$ dynamic template are used, and supervision is provided by predicting the target on the dynamic template. 
In the second stage, the full configuration is enabled, in which one initial template, two dynamic templates, and one $378\times 378$ search region are jointly fed into the network, and supervision is computed on the search region. 
The initial template, the dynamic templates, and the search region are each routed through a dedicated LoRA branch. Both stages use the AdamW~\cite{DBLP:conf/iclr/LoshchilovH19_AdamW} optimizer with an initial learning rate of $10^{-7}$, a one-epoch warmup to $10^{-4}$, a cosine schedule decaying to $5\times 10^{-6}$, and a weight decay of $0.1$.


\noindent\textbf{Loss and Inference of ACTrack-EM.}
For target prediction, we employ Generalized IoU \cite{rezatofighi2019generalized} Loss to supervise bounding box prediction, and Binary Cross-Entropy Loss to supervise target center point prediction. Additionally, we use Cross-Entropy Loss to compute the modality prediction loss. The loss weights for the above components are all set to 1.0.
For the two auxiliary templates in ACTrack-EM, an update is performed only when the center-point confidence produced by the tracker exceeds $0.9$ and the gap to the previous update is at least $\lfloor n/5 \rfloor$ frames, where $n$ denotes the number of frames already tracked. When both conditions are satisfied, the auxiliary templates are refreshed in a FIFO manner.

\subsection{Comparison with the State-of-the-Art Methods}

We evaluate the proposed ACTrack-B$_{224}$, ACTrack-L$_{384}$, and ACTrack-EM on twelve datasets spanning five modalities: RGB, RGB-D, RGB-E, RGB-T, and RGB-Language. For the RGB modality, we primarily report ACTrack-B$_{224}$ and ACTrack-L$_{384}$, whereas for the remaining multimodal benchmarks we primarily report ACTrack-EM. We compare against recent state-of-the-art trackers and highlight the best three results for each metric in \textbf{\textcolor{red}{red}}, \textbf{\textcolor{blue}{blue}}, and \textbf{bold}.

\begin{figure}[!t]
\centering
\includegraphics[width=\linewidth]{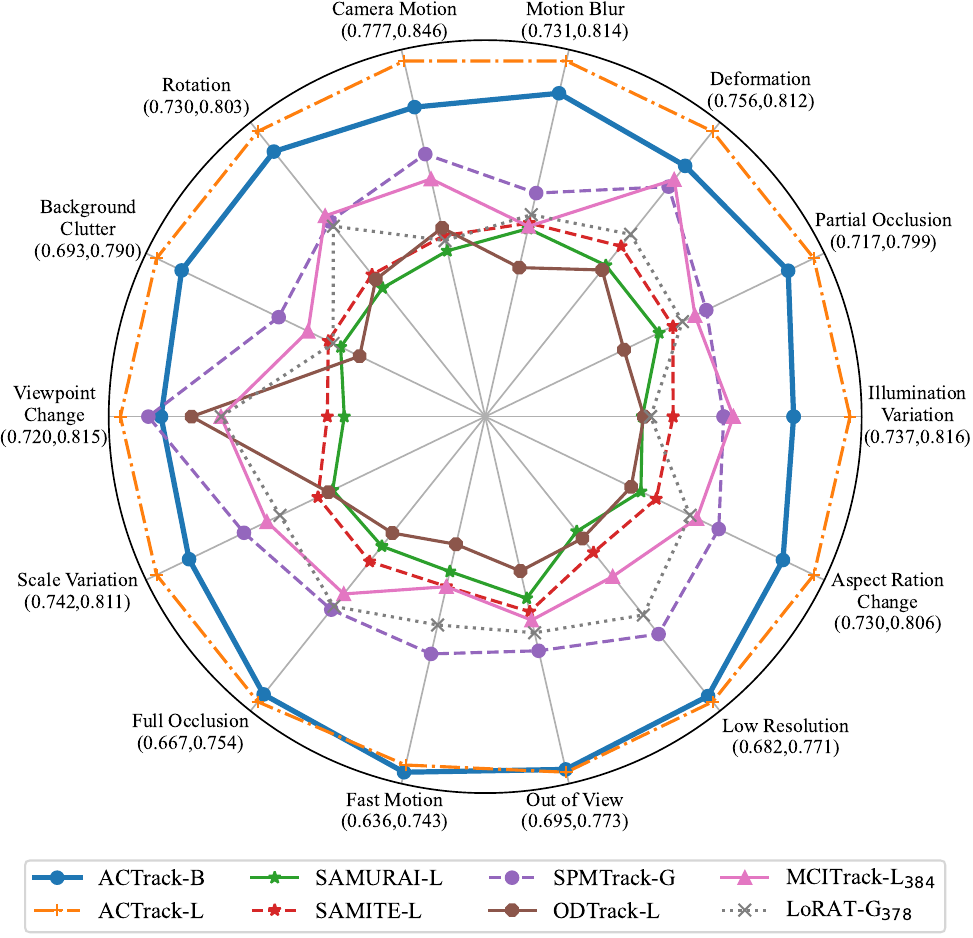}
\caption{The performance of our method compared with other state-of-the-art trackers in terms of AUC across various scenarios in the LaSOT \emph{test} split.}
\label{fig:lasotSubset}
\end{figure}

\noindent\textbf{LaSOT.}
LaSOT~\cite{fan2019lasot} is an \textbf{RGB-based} tracking dataset built for long-term tracking, whose test split contains 280 sequences averaging more than 2{,}500 frames. 
As shown in Table~\ref{tab:rgb_lasot_trackingnet}, our proposed ACTrack-B$_{224}$ attains an AUC of $80.1$, substantially outperforming all existing methods and being the first to push the AUC beyond $80$. 
ACTrack-L$_{384}$ further achieves the best AUC of $81.4$.
In addition, Fig.~\ref{fig:lasotSubset} compares ACTrack with other methods on each attribute-based subset of LaSOT, where our method significantly outperforms all competitors across every subset.

\noindent\textbf{LaSOT$_\mathrm{ext}$.}
LaSOT$_\mathrm{ext}$~\cite{fan2019lasot} is an \textbf{RGB-based} extension that adds 150 sequences of 15 categories disjoint from LaSOT, probing generalization to unseen classes. 
As shown in Table~\ref{tab:rgb_lasot_trackingnet}, similar to the results on LaSOT, our method also achieves a substantial performance gain, both ACTrack variants rank first and second across all three metrics.

\noindent\textbf{TrackingNet.}
TrackingNet~\cite{muller2018trackingnet} is an \textbf{RGB-based} large-scale short-term dataset with 511 test sequences. As shown in Table~\ref{tab:rgb_lasot_trackingnet}, ACTrack-L$_{384}$ reaches the best AUC of $88.0$, showing that our method also preserves strong short-term localization. As performance on TrackingNet has nearly saturated, the gain over prior methods is less pronounced than on the other datasets.

\begin{table}[!h]
\centering
\caption{The performance of our method and other state-of-the-art trackers on GOT-10k. The best three results for each metric are highlighted in \textbf{\textcolor{red}{red}}, \textbf{\textcolor{blue}{blue}}, and \textbf{bold}, respectively.}
\label{tab:got10k}
\footnotesize
\setlength{\tabcolsep}{3.5pt}
\begin{tabular*}{\columnwidth}{@{\extracolsep{\fill}}l|c|ccc}
\toprule
\textbf{Method} & \textbf{Source} & \textbf{AO} & $\mathbf{SR_{0.5}}$ & $\mathbf{SR_{0.75}}$ \\
\midrule
\textbf{ACTrack-B$_{224}$} & Ours & \second{83.8} & \second{95.4} & \second{82.0} \\
\textbf{ACTrack-L$_{384}$} & Ours & \first{86.8} & \first{95.9} & \first{87.2} \\
\midrule
ARPTrack-L$_{384}$~\cite{Liang_2025_CVPR_arptrack} & CVPR25 & \third{81.5} & \third{90.6} & 80.5 \\
SPMTrack-L~\cite{Cai_2025_CVPR_SPMTrack} & CVPR25 & 80.0 & 89.4 & 79.9 \\
MCITrack-L$_{384}$~\cite{kang2025exploring_mcitrack} & AAAI25 & 80.0 & 88.5 & 80.2 \\
ARTrackV2-L$_{384}$~\cite{Bai_2024_CVPR_ARTrackV2} & CVPR24 & 79.5 & 87.8 & 79.6 \\
LoRAT-g$_{378}$~\cite{LoRAT} & ECCV24 & 78.9 & 87.8 & \third{80.7} \\
LoRATv2-L$_{378}$~\cite{lin2025lorat_v2} & NeurIPS25 & 78.2 & 86.8 & 79.1 \\
HIPTrack~\cite{Cai_2024_CVPR_HIPTrack} & CVPR24 & 77.4 & 88.0 & 74.5 \\
AQATrack$_{384}$~\cite{Xie_2024_CVPR_AQATrack} & CVPR24 & 76.0 & 85.2 & 74.9 \\
MambaLCT$_{384}$~\cite{li2025mambalct} & AAAI25 & 76.2 & 86.7 & 74.3 \\
ARTrack-L$_{384}$~\cite{Wei_2023_CVPR_autoregressive} & CVPR23 & 78.5 & 87.4 & 77.8 \\
ODTrack-L$_{384}$~\cite{ODTrack} & AAAI24 & 78.2 & 87.2 & 77.3 \\
LoRAT-L$_{378}$~\cite{LoRAT} & ECCV24 & 77.5 & 86.2 & 78.1 \\
SeqTrack-L$_{384}$~\cite{Chen_2023_CVPR_seqtrack} & CVPR23 & 74.8 & 81.9 & 72.2 \\
\bottomrule
\end{tabular*}
\end{table}

\noindent\textbf{GOT-10k.}
GOT-10k~\cite{huang2019got} is an \textbf{RGB-based} dataset with 9{,}335 training and 180 test sequences under a strict one-shot protocol, where test classes do not overlap with the training split. As shown in Table~\ref{tab:got10k}, ACTrack-L$_{384}$ obtains the best AO of $86.8$. Since current state-of-the-art methods commonly rely on large-scale pre-trained foundation models, the GOT-10k evaluation may no longer strictly conform to its one-shot protocol; we therefore report GOT-10k separately for reference.

\noindent\textbf{TNL2K.}
TNL2K~\cite{Wang_2021_CVPR_TNL2k} is an \textbf{RGB-language} dataset whose videos are annotated with natural-language descriptions under complex scenes and severe distractors. As shown in Table~\ref{tab:tnl2k}, our proposed ACTrack-EM achieves the best AUC of $77.2$, outperforming existing methods by a clear margin.

\begin{table}[!h]
\centering
\caption{The performance of our method and other state-of-the-art trackers on RGB-Language dataset TNL2K. We report AUC, normalized precision, and precision when available. Entries marked ``--'' are not reported in the corresponding paper or public benchmark table. The best three results for each metric are highlighted in \textbf{\textcolor{red}{red}}, \textbf{\textcolor{blue}{blue}}, and \textbf{bold}, respectively.}
\label{tab:tnl2k}
\footnotesize
\setlength{\tabcolsep}{2.6pt}
\begin{tabular*}{\columnwidth}{@{\extracolsep{\fill}}l|c|ccc}
\toprule
\textbf{Method} & \textbf{Source} & \textbf{AUC} & $\mathbf{P_\mathrm{Norm}}$ & $\mathbf{P}$ \\
\midrule
\textbf{ACTrack-EM} & Ours & \first{77.2} & \first{87.1} & \first{84.7} \\
\midrule
SUTrack-L$_{384}$~\cite{sutrack} & AAAI25 & \second{67.9} & -- & \second{72.1} \\
MCITrack-L$_{384}$~\cite{kang2025exploring_mcitrack} & AAAI25 & \third{65.3} & -- & -- \\
UVLTrack-L~\cite{ma2024uvltrack} & AAAI24 & 64.8 & \second{82.8} & 68.8 \\
SPMTrack-G~\cite{Cai_2025_CVPR_SPMTrack} & CVPR25 & 64.7 & \third{82.6} & \third{70.6} \\
RELO-L$_{256}$~\cite{chen2026reloreinforcementlearninglocalize} & ICML26 & 63.6 & -- & -- \\
LoRATv2-L$_{378}$~\cite{lin2025lorat_v2} & NeurIPS25 & 62.4 & -- & 67.7 \\
LoRAT-g$_{378}$~\cite{LoRAT} & ECCV24 & 62.7 & -- & 67.8 \\
ODTrack-L~\cite{ODTrack} & AAAI24 & 61.7 & -- & -- \\
ARTrackV2-L$_{384}$~\cite{Bai_2024_CVPR_ARTrackV2} & CVPR24 & 61.6 & -- & -- \\
RTracker-L~\cite{Huang_2024_CVPR_RTracker} & CVPR24 & 60.6 & -- & 63.7 \\
OneTracker~\cite{Hong_2024_CVPR_onetracker} & CVPR24 & 58.0 & -- & 59.1 \\
JointNLT~\cite{Zhou_2023_CVPR_jointnlt} & CVPR23 & 56.9 & 73.5 & 58.1 \\
\bottomrule
\end{tabular*}
\end{table}

\noindent\textbf{OTB2015, UAV123, and NfS.}
OTB2015~\cite{otb2015}, UAV123~\cite{mueller2016benchmark}, and NfS~\cite{Galoogahi_2017_ICCV_nfs} are classical \textbf{RGB-based} short-term datasets.
OTB2015 contains 100 sequences with 11 challenge attributes, UAV123 contains 123 low-altitude aerial sequences with small fast-moving targets, and NfS contains sequences for evaluating tracking under fast motion. 
For NfS, we use the 30 FPS version.
As shown in Table~\ref{tab:otb_uav_nfs}, ACTrack-B$_{224}$ achieves the best AUC on all three benchmarks, surpassing all previous large and giant scale methods.

\begin{table}[!h]
\centering
\caption{The performance of our method and other state-of-the-art trackers on OTB2015, UAV123, and NfS (30~fps). We report AUC metric. Entries marked ``--'' are not reported in the corresponding paper. The best three results for each metric are highlighted in \textbf{\textcolor{red}{red}}, \textbf{\textcolor{blue}{blue}}, and \textbf{bold}, respectively.}
\label{tab:otb_uav_nfs}
\footnotesize
\setlength{\tabcolsep}{3pt}
\begin{tabular*}{\columnwidth}{@{\extracolsep{\fill}}l|c|ccc}
\toprule
\textbf{Method} & \textbf{Source} & \textbf{OTB2015} & \textbf{UAV123} & \textbf{NfS} \\
\midrule
\textbf{ACTrack-B$_{224}$} & Ours & \first{73.9} & \first{74.3} & \first{73.7} \\
\midrule
RELO-L$_{256}$~\cite{chen2026reloreinforcementlearninglocalize} & ICML26 & -- & 71.4 & \second{71.3} \\
MCITrack-L$_{384}$~\cite{kang2025exploring_mcitrack} & AAAI25 & -- & 71.5 & \third{70.6} \\
LoRAT-g$_{378}$~\cite{LoRAT} & ECCV24 & \second{72.6} & \second{73.9} & 68.1 \\
ARTrackV2-L$_{384}$~\cite{Bai_2024_CVPR_ARTrackV2} & CVPR24 & -- & \third{71.7} & 68.4 \\
HIPTrack~\cite{Cai_2024_CVPR_HIPTrack} & CVPR24 & \third{71.0} & 70.5 & 68.1 \\
ARTrack-L$_{384}$~\cite{Wei_2023_CVPR_autoregressive} & CVPR23 & -- & 71.2 & 67.9 \\
SeqTrack-L$_{384}$~\cite{Chen_2023_CVPR_seqtrack} & CVPR23 & -- & 68.5 & 66.2 \\
AiATrack~\cite{gao2022aiatrack} & ECCV22 & 69.6 & 70.6 & 67.9 \\
MixFormer-L~\cite{Cui_2022_MixFormer} & CVPR22 & -- & 69.5 & -- \\
KeepTrack~\cite{mayer2021learning} & ICCV21 & 70.9 & 69.7 & 66.4 \\
TransT~\cite{chen2021transformer} & CVPR21 & 69.4 & 69.1 & 65.7 \\
\bottomrule
\end{tabular*}
\end{table}



\noindent\textbf{VastTrack.}
VastTrack~\cite{NEURIPS2024_ec17a52e_vasttrack} is a recent large-scale \textbf{RGB-based} dataset whose test split spans hundreds of object categories, making it substantially more challenging than LaSOT. 
As shown in Table~\ref{tab:vasttrack}, all three ACTrack variants surpass all competing trackers by a large margin, demonstrating strong category-level generalization.

\begin{table}
\centering
\caption{Comparison on VastTrack. We report AUC and precision (P). Numbers are taken from the public comparison tables of VastTrack~\cite{NEURIPS2024_ec17a52e_vasttrack} and LoRATv2~\cite{lin2025lorat_v2}. The best three results for each metric are highlighted in \textbf{\textcolor{red}{red}}, \textbf{\textcolor{blue}{blue}}, and \textbf{bold}, respectively.}
\label{tab:vasttrack}
\footnotesize
\setlength{\tabcolsep}{4.5pt}
\begin{tabular*}{\columnwidth}{@{\extracolsep{\fill}}l|c|cc}
\toprule
\textbf{Method} & \textbf{Source} & \textbf{AUC} & $\mathbf{P}$ \\
\midrule
\textbf{ACTrack-L$_{384}$} & Ours & \second{69.0} & \second{77.8} \\
\textbf{ACTrack-B$_{224}$} & Ours & \third{65.9} & \third{73.2} \\
\textbf{ACTrack-EM} & Ours & \first{69.9} & \first{79.0} \\
\midrule
LoRATv2-L$_{378}$~\cite{lin2025lorat_v2} & NeurIPS25 & 44.2 & 46.7 \\
LoRAT-L$_{378}$~\cite{LoRAT} & ECCV24 & 43.9 & 45.8 \\
SeqTrack-L$_{384}$~\cite{Chen_2023_CVPR_seqtrack} & CVPR23 & 39.6 & 40.2 \\
MixFormer-L~\cite{Cui_2022_MixFormer} & CVPR22 & 39.5 & 39.8 \\
ROMTrack$_{384}$~\cite{Cai_2023_ICCV_ROMTrack} & ICCV23 & 37.0 & 36.1 \\
DropTrack~\cite{Wu_2023_CVPR_dropmae} & CVPR23 & 37.0 & 36.5 \\
GRM~\cite{Gao_2023_CVPR_GRM} & CVPR23 & 36.3 & 34.8 \\
ARTrack-L$_{384}$~\cite{Wei_2023_CVPR_autoregressive} & CVPR23 & 35.6 & 32.4 \\
OSTrack$_{384}$~\cite{ye_2022_joint} & ECCV22 & 33.6 & 31.5 \\
STARK~\cite{yan2021learning} & ICCV21 & 33.4 & 30.8 \\
TransT~\cite{chen2021transformer} & CVPR21 & 29.9 & 25.4 \\
\bottomrule
\end{tabular*}
\end{table}

\begin{table*}[t]
\centering
\caption{State-of-the-art comparison on multimodal tracking benchmarks. LasHeR evaluates RGB-T tracking, VisEvent evaluates RGB-Event tracking, and DepthTrack evaluates RGB-D tracking. The best three results for each metric are highlighted in \textbf{\textcolor{red}{red}}, \textbf{\textcolor{blue}{blue}}, and \textbf{bold}, respectively.}
\label{tab:multimodal_main}
\footnotesize
\setlength{\tabcolsep}{3.2pt}
\renewcommand{\arraystretch}{1.05}
\begin{tabular*}{\textwidth}{@{\extracolsep{\fill}}l|c|cc|cc|ccc}
\toprule
\textbf{Method} & \textbf{Source} & \multicolumn{2}{c|}{\textbf{LasHeR}} & \multicolumn{2}{c|}{\textbf{VisEvent}} & \multicolumn{3}{c}{\textbf{DepthTrack}} \\
\cmidrule(lr){3-4}\cmidrule(lr){5-6}\cmidrule(lr){7-9}
 & & \textbf{SR} & \textbf{PR} & \textbf{AUC} & $\mathbf{P}$ & \textbf{F-score} & \textbf{Re} & \textbf{Pr} \\
\midrule
\textbf{ACTrack-EM} & Ours & \first{63.7} & \first{79.4} & \first{77.8} & \first{94.6} & \first{74.9} & \first{76.5} & \first{73.3} \\
\midrule
Uni-MDTrack-L~\cite{cai2026unimdtracklearningdecoupledmemory} & arXiv26 & \second{62.1} & \second{77.9} & \second{65.7} & \second{81.8} & \second{67.4} & \second{67.2} & \second{67.6} \\
FlexTrack~\cite{tan2025you_flextrack} & ICCV25 & \third{62.0} & \third{77.3} & 64.1 & \third{81.4} & \third{67.0} & \third{66.9} & \third{67.1} \\
Uni-MDTrack-B~\cite{cai2026unimdtracklearningdecoupledmemory} & arXiv26 & 61.2 & 76.7 & \third{64.2} & 81.0 & 65.9 & 66.3 & 66.2 \\
SUTrack-L$_{224}$~\cite{sutrack} & AAAI25 & 61.9 & 77.0 & 64.0 & 80.9 & 64.3 & 64.6 & 64.0 \\
SUTrack-L$_{384}$~\cite{sutrack} & AAAI25 & 61.9 & 76.9 & 63.8 & 80.5 & 66.4 & 66.4 & 66.5 \\
STTrack~\cite{hu2025exploiting_sttrack} & AAAI25 & 60.3 & 76.0 & 61.9 & 78.6 & 63.3 & 63.4 & 63.2 \\
SeqTrackV2-L$_{384}$~\cite{chen2023unified_seqtrackv2} & arXiv23 & 61.0 & 76.7 & 63.4 & -- & 62.3 & 62.6 & 62.5 \\
OneTracker~\cite{Hong_2024_CVPR_onetracker} & CVPR24 & 53.8 & 67.2 & 60.8 & 76.7 & 60.9 & 60.4 & 60.7 \\
UnTrack~\cite{Wu_2024_CVPR_untrack} & CVPR24 & 53.6 & 66.7 & 58.9 & 75.5 & 61.2 & 61.0 & 61.3 \\
SDSTrack~\cite{Hou_2024_CVPR_sdstrack} & CVPR24 & 53.1 & 66.5 & 59.7 & 76.7 & 61.4 & 60.9 & 61.9 \\
ViPT~\cite{Zhu_2023_CVPR_vipt} & CVPR23 & 52.5 & 65.1 & 59.2 & 75.8 & 59.4 & 59.6 & 59.2 \\
ProTrack~\cite{yang2022prompting_protrack} & MM22 & 42.0 & 53.8 & 47.1 & 63.2 & 57.8 & 57.3 & 58.3 \\
\bottomrule
\end{tabular*}
\end{table*}


\noindent\textbf{LasHeR.}
LasHeR~\cite{9640453_lasher} is a large-scale \textbf{RGB-Thermal} dataset with aligned visible and thermal sequences. As shown in Table~\ref{tab:multimodal_main}, ACTrack-EM achieves the best SR and PR.

\noindent\textbf{VisEvent.}
VisEvent~\cite{10284004_visevent} is an \textbf{RGB-Event} dataset pairing frames with event-camera streams. 
As shown in Table~\ref{tab:multimodal_main}, ACTrack-EM reaches state-of-the-art performance, improving upon the previous best result by more than $10\%$ AUC and thus outperforming all other methods by a large margin.

\noindent\textbf{DepthTrack.}
DepthTrack~\cite{Yan_2021_ICCV_depthtrack} is a long-term \textbf{RGB-Depth} dataset reported with F-score, recall (Re), and precision (Pr). As shown in Table~\ref{tab:multimodal_main}, ACTrack-EM obtains the best results across all three metrics.

\begin{figure*}[!t]
\centering
\begin{minipage}{0.193\textwidth}
\centering
\includegraphics[width=\linewidth]{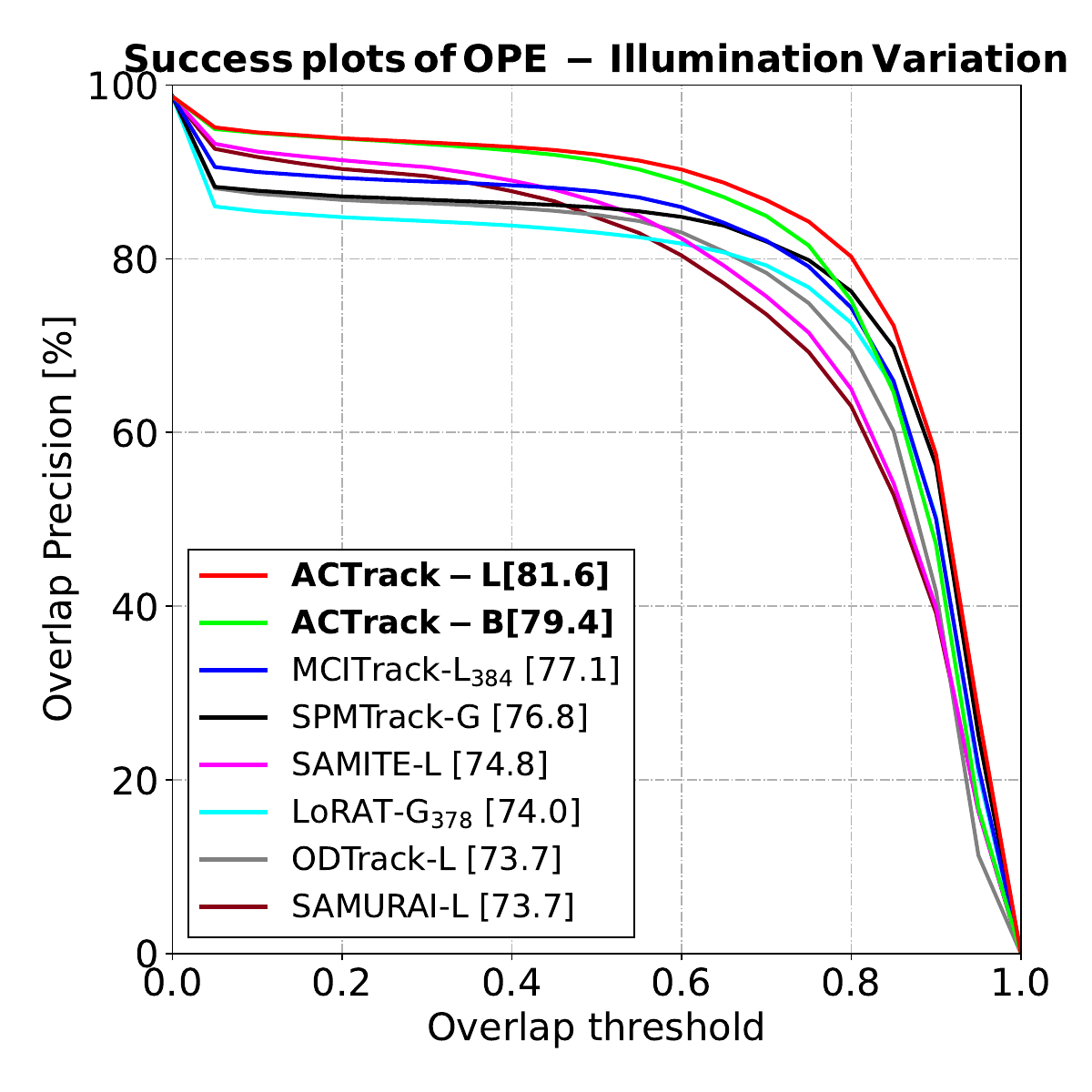}
\end{minipage}\hfill
\begin{minipage}{0.193\textwidth}
\centering
\includegraphics[width=\linewidth]{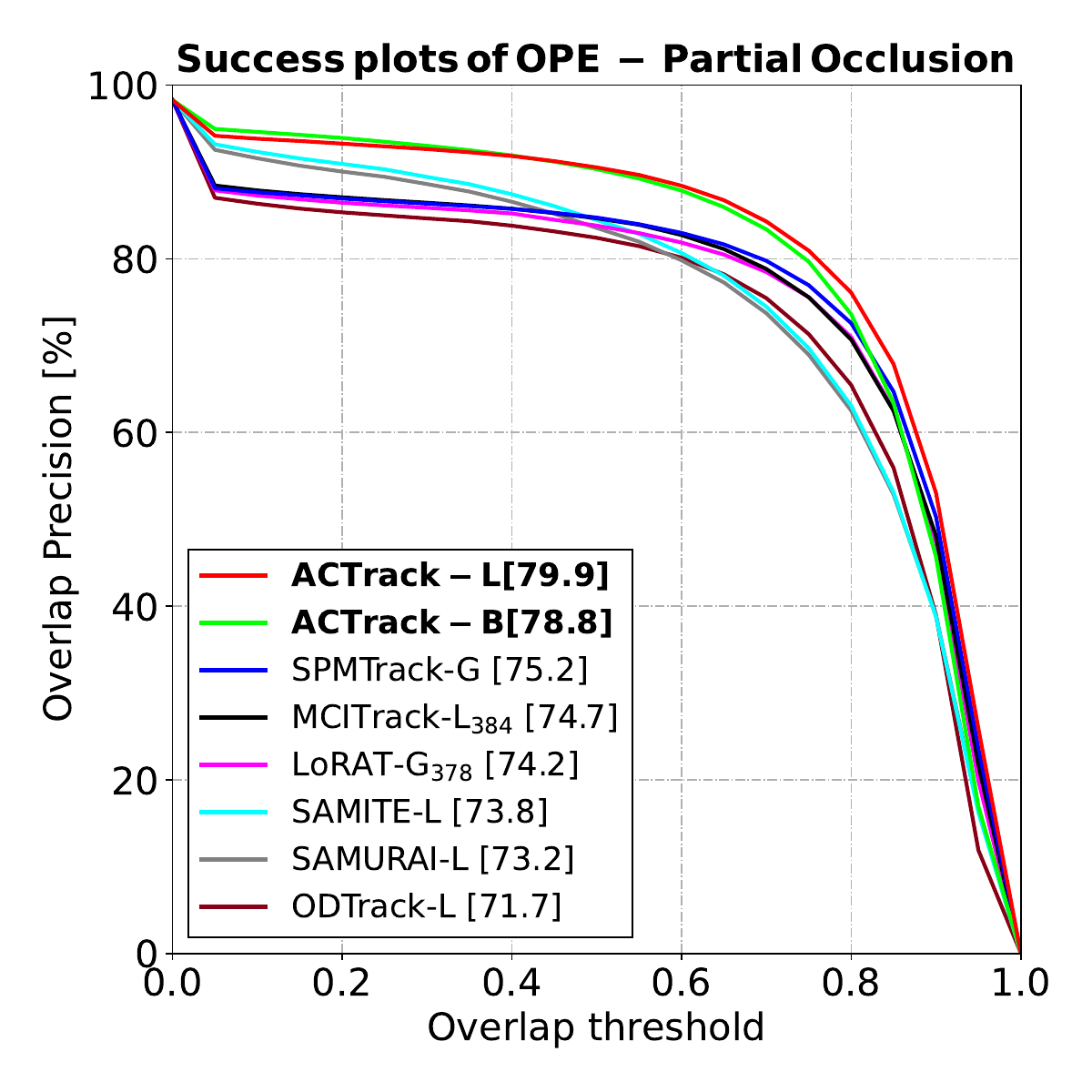}
\end{minipage}\hfill
\begin{minipage}{0.193\textwidth}
\centering
\includegraphics[width=\linewidth]{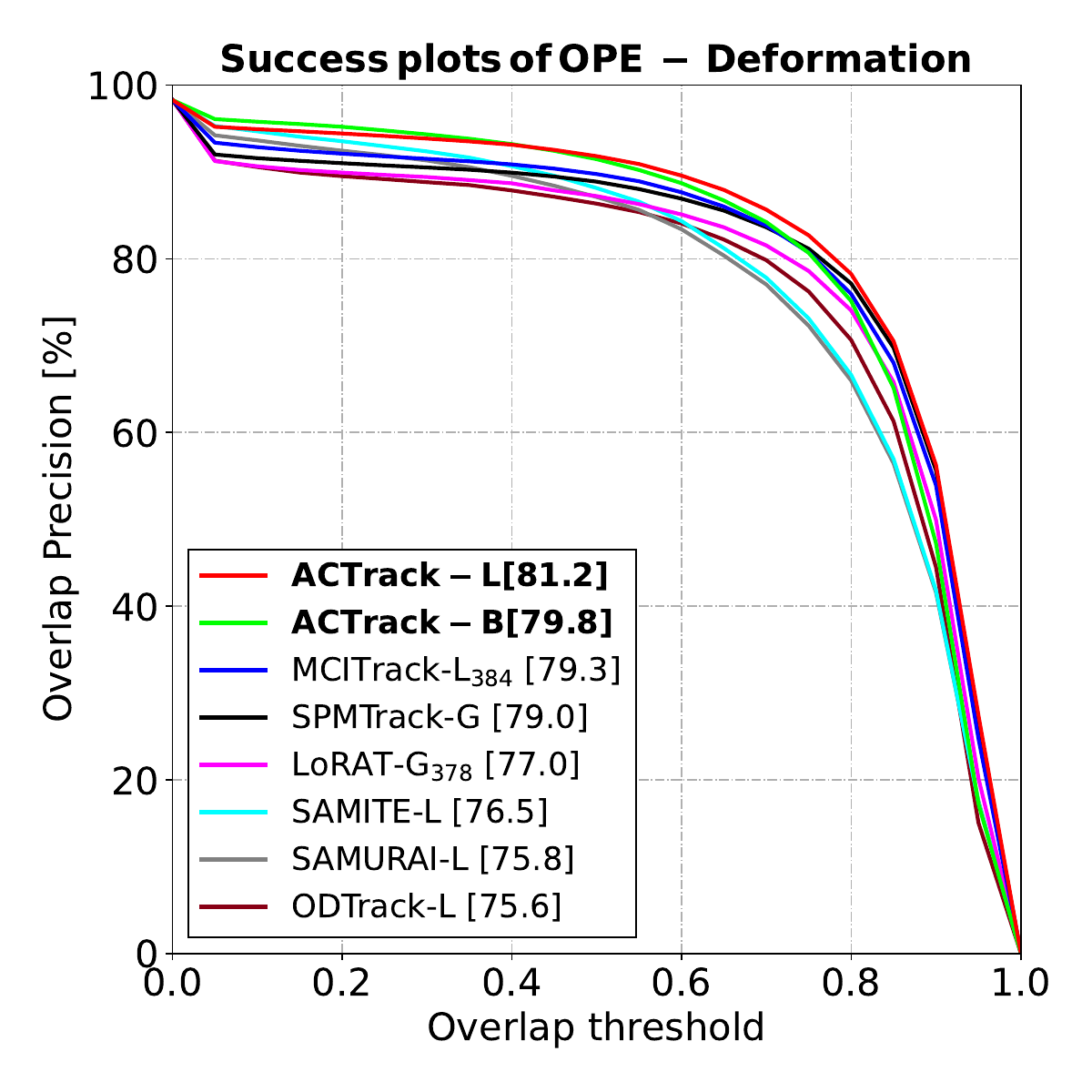}
\end{minipage}\hfill
\begin{minipage}{0.193\textwidth}
\centering
\includegraphics[width=\linewidth]{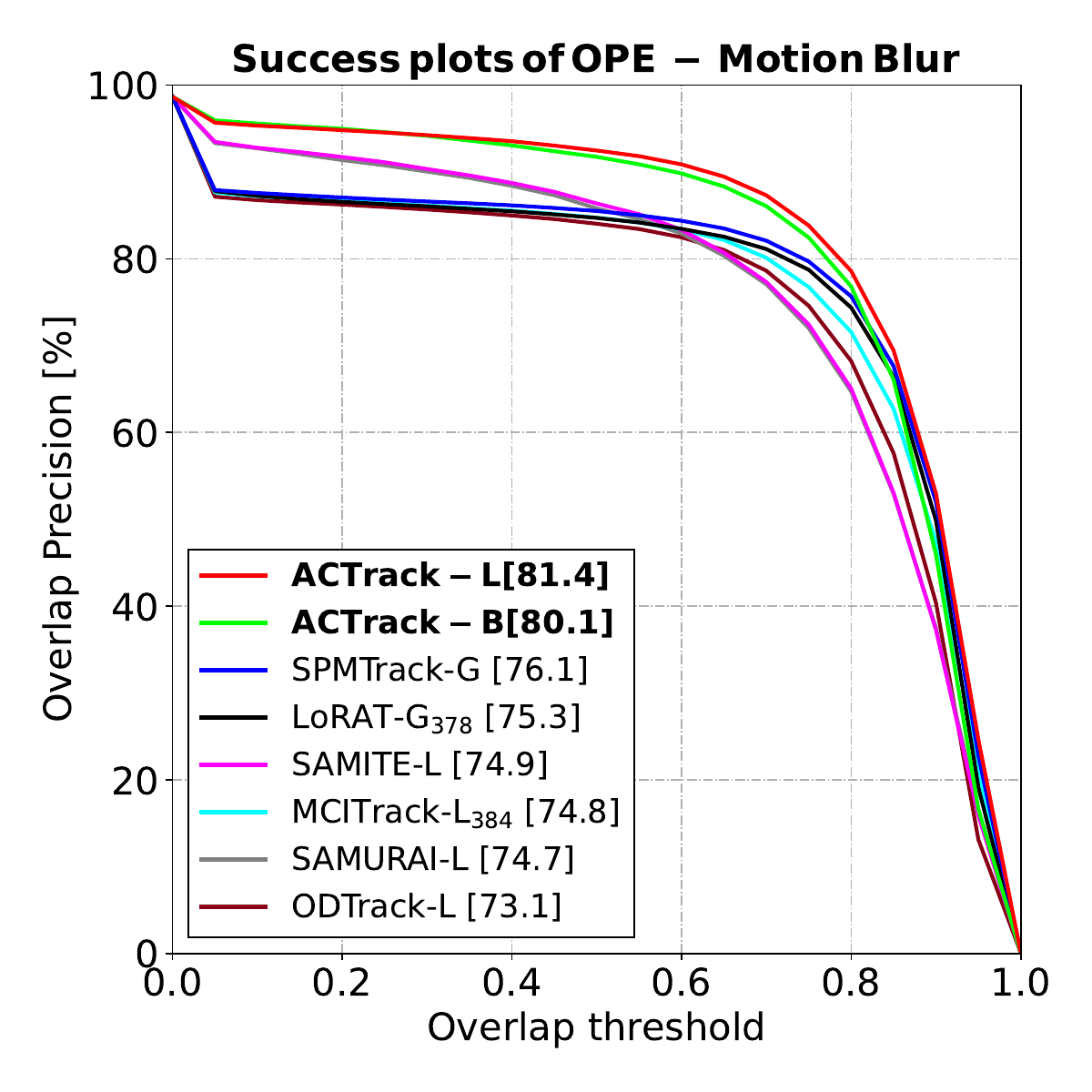}
\end{minipage}\hfill
\begin{minipage}{0.193\textwidth}
\centering
\includegraphics[width=\linewidth]{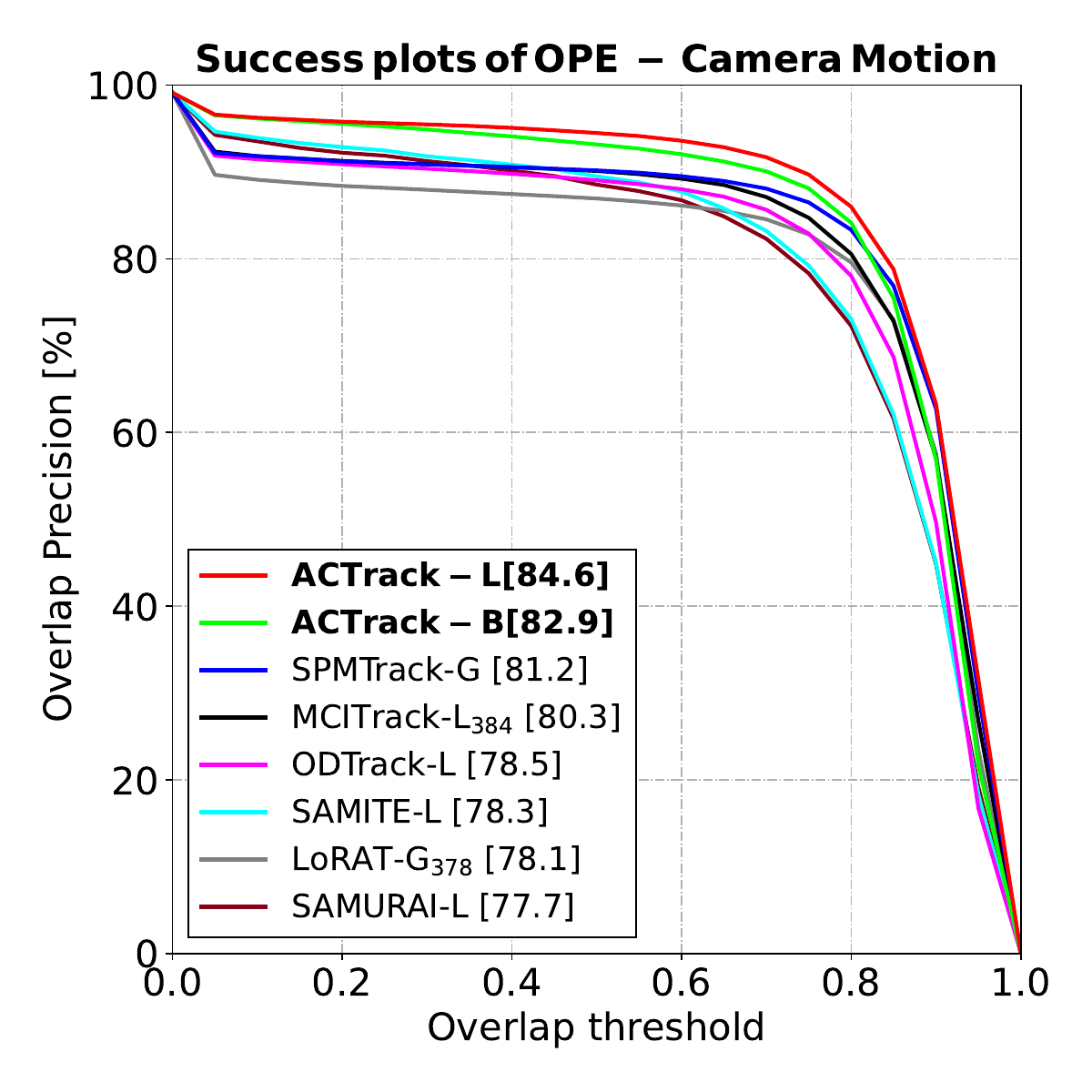}
\end{minipage}

\vspace{0.6ex}

\begin{minipage}{0.193\textwidth}
\centering
\includegraphics[width=\linewidth]{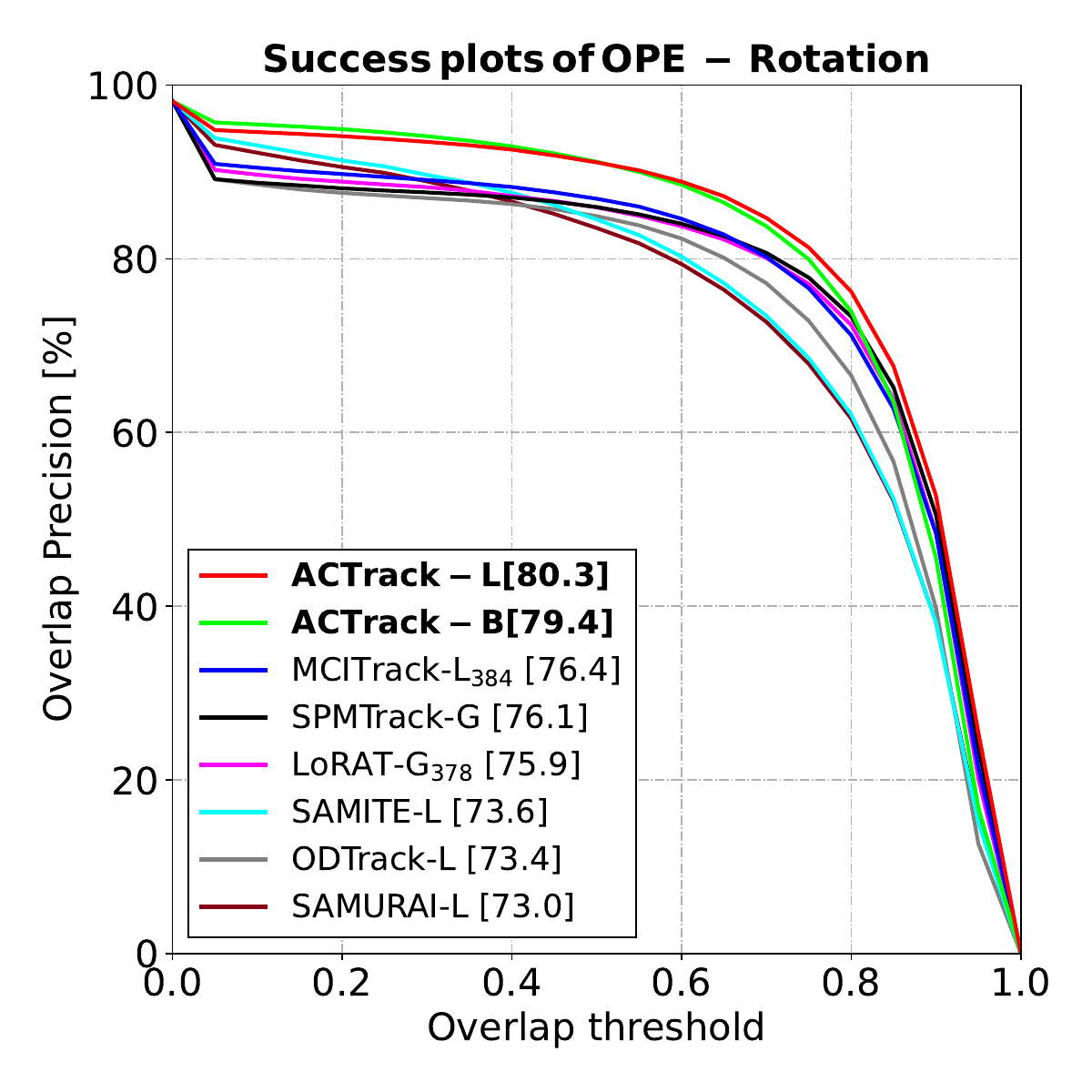}
\end{minipage}\hfill
\begin{minipage}{0.193\textwidth}
\centering
\includegraphics[width=\linewidth]{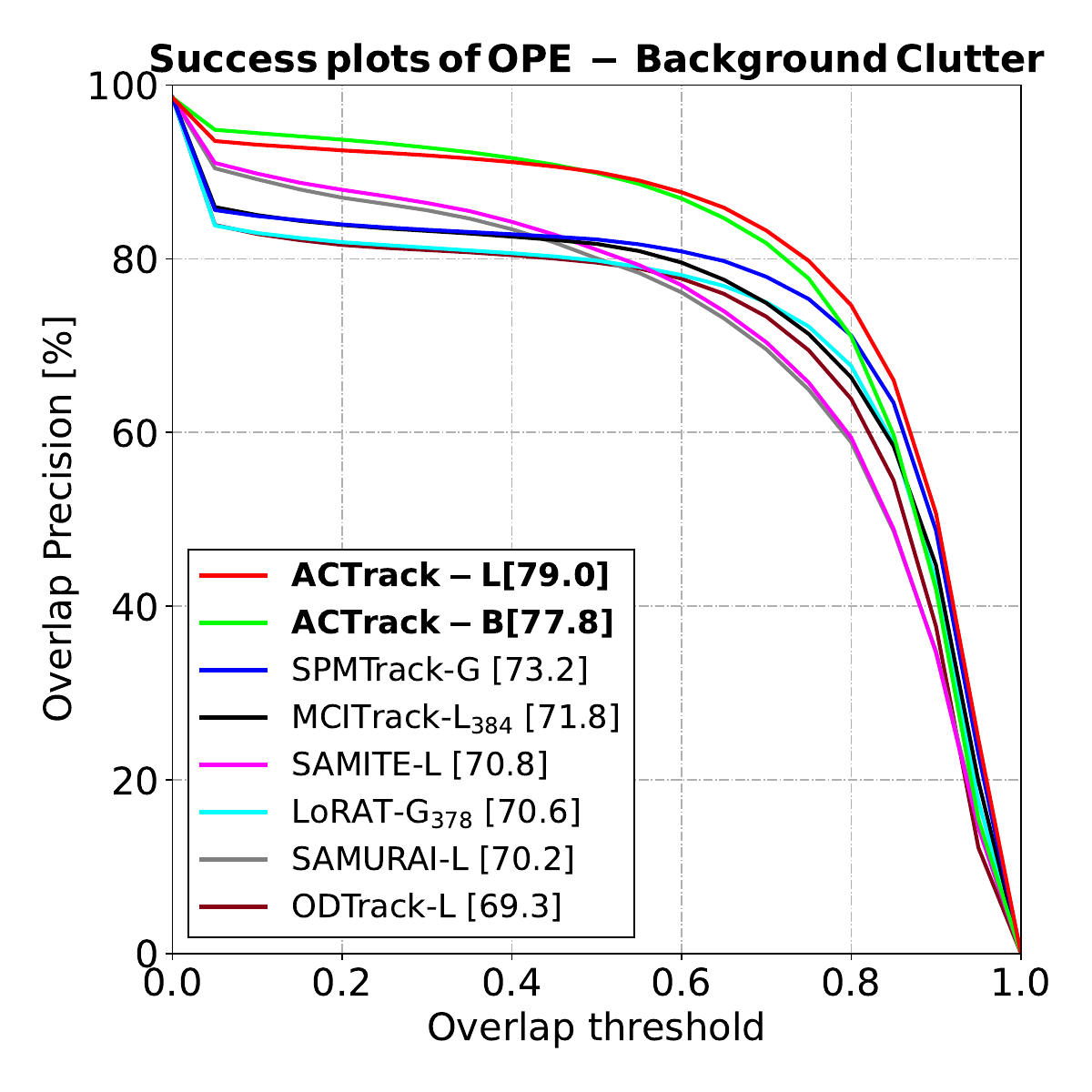}
\end{minipage}\hfill
\begin{minipage}{0.193\textwidth}
\centering
\includegraphics[width=\linewidth]{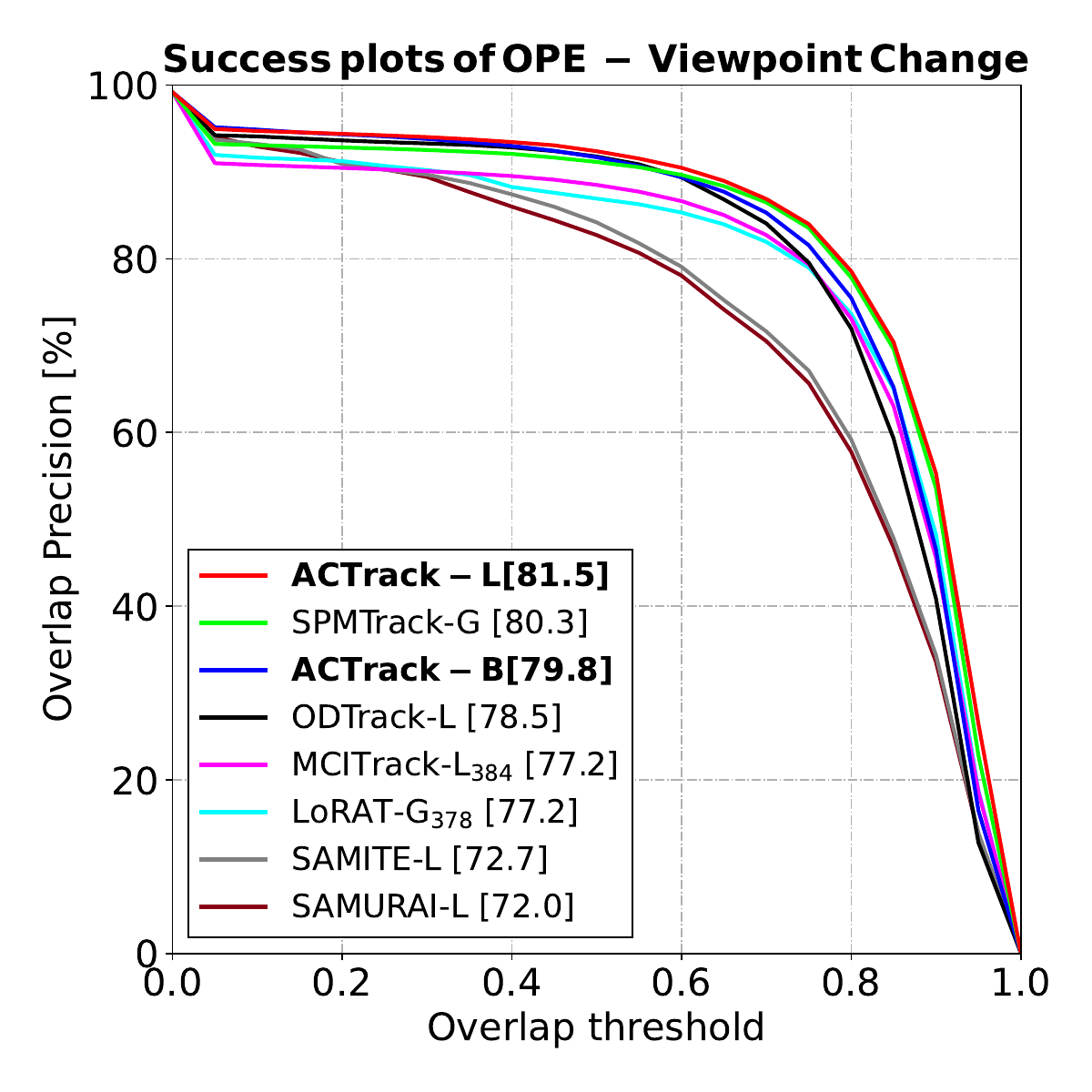}
\end{minipage}\hfill
\begin{minipage}{0.193\textwidth}
\centering
\includegraphics[width=\linewidth]{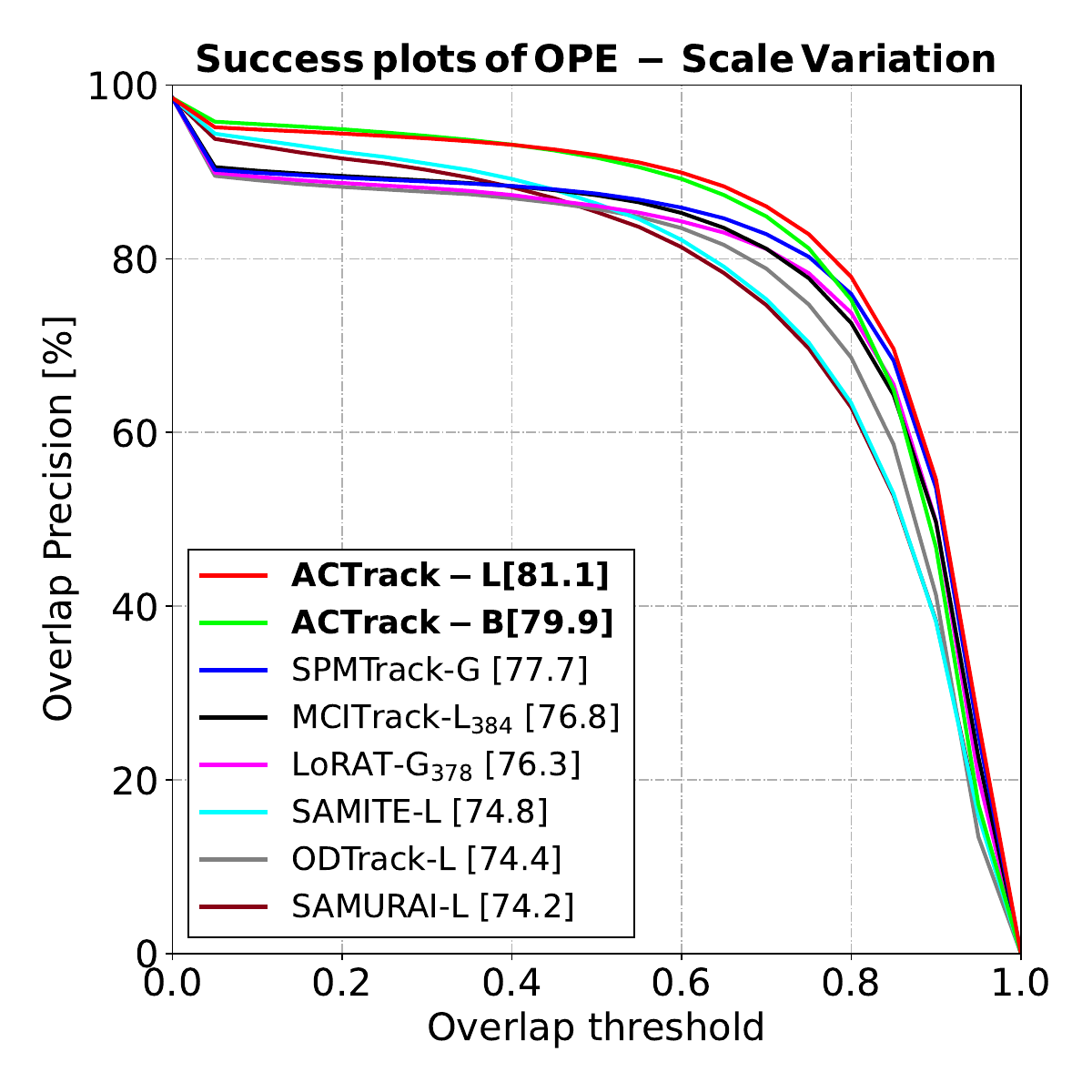}
\end{minipage}\hfill
\begin{minipage}{0.193\textwidth}
\centering
\includegraphics[width=\linewidth]{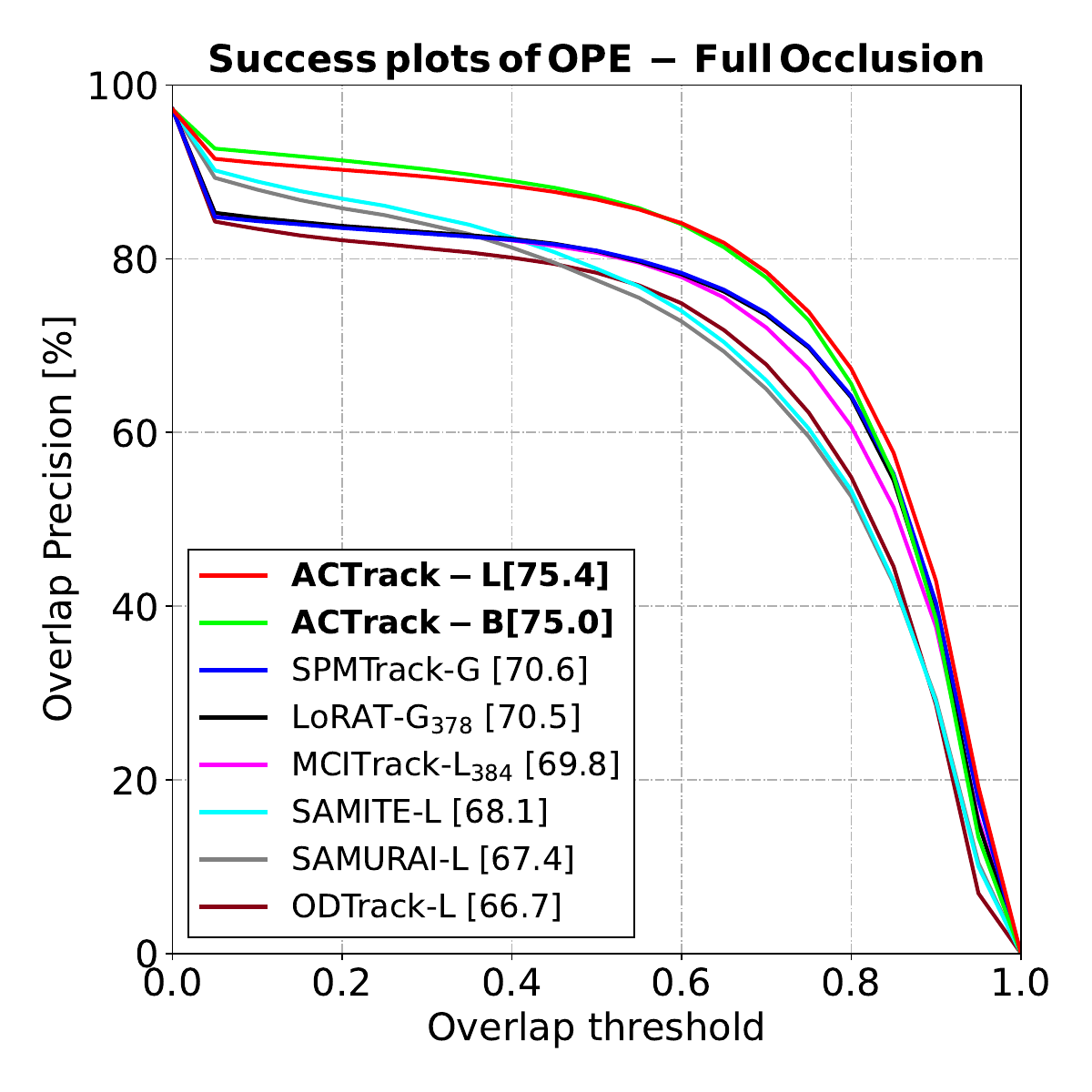}
\end{minipage}

\vspace{0.6ex}

\begin{minipage}{0.193\textwidth}
\centering
\includegraphics[width=\linewidth]{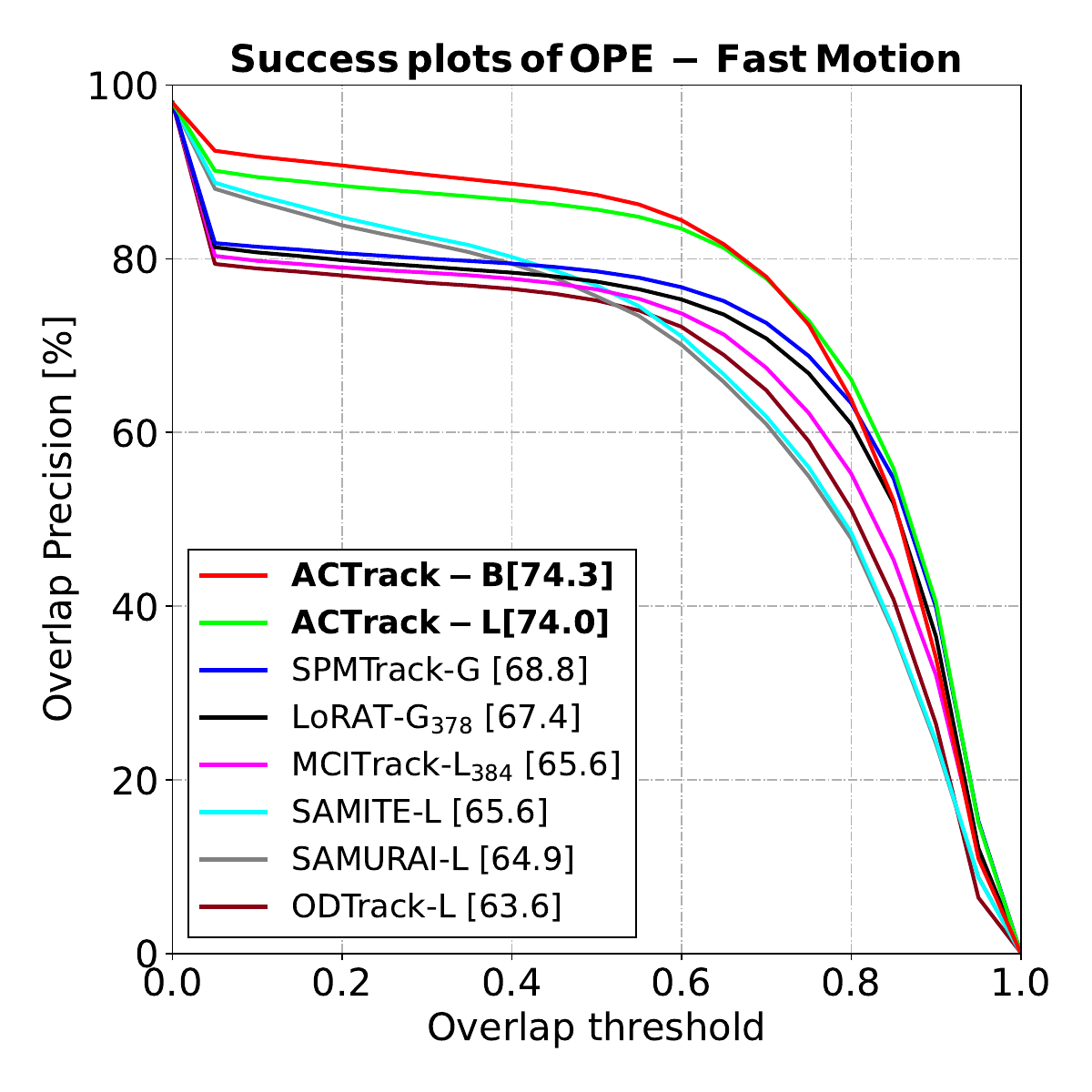}
\end{minipage}\hfill
\begin{minipage}{0.193\textwidth}
\centering
\includegraphics[width=\linewidth]{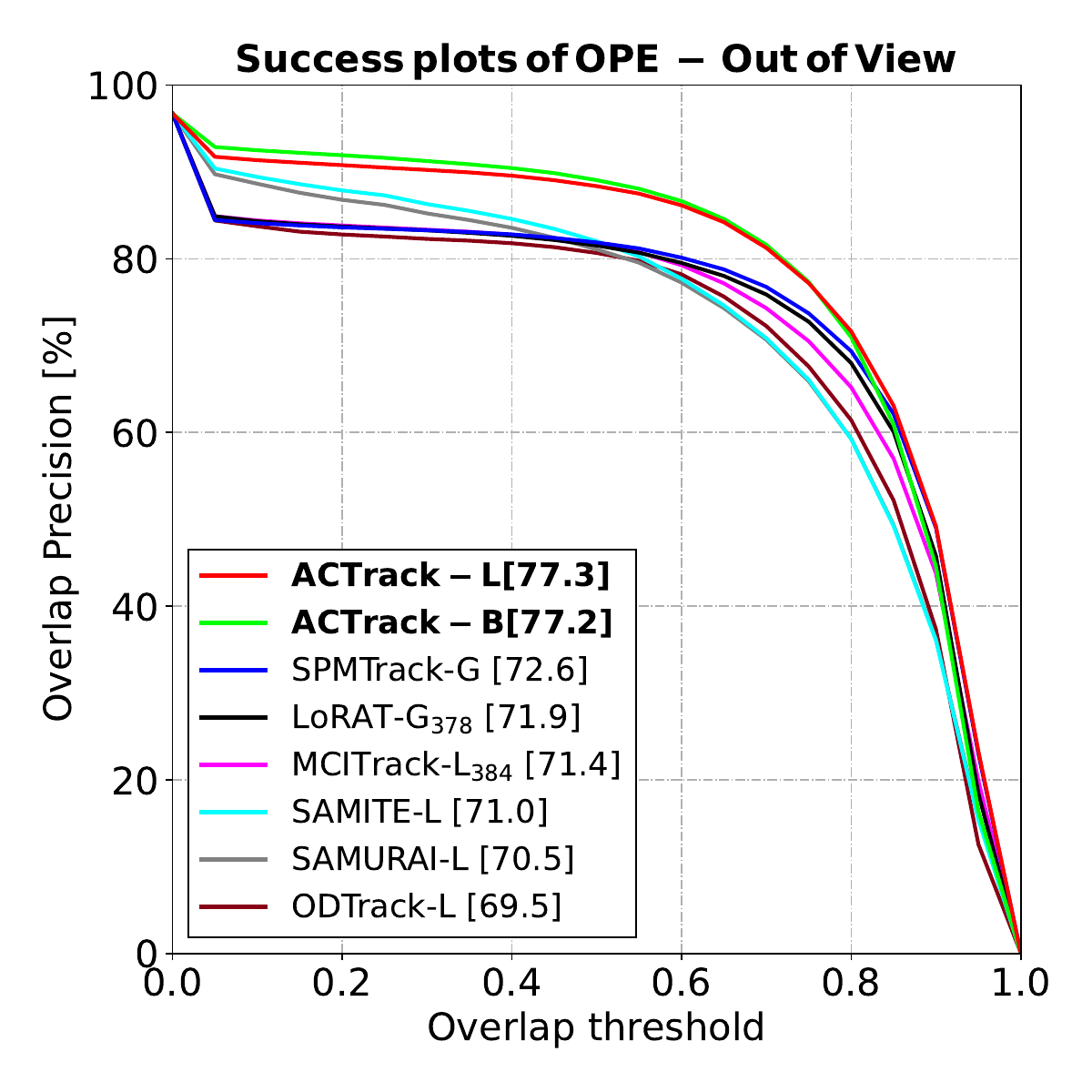}
\end{minipage}\hfill
\begin{minipage}{0.193\textwidth}
\centering
\includegraphics[width=\linewidth]{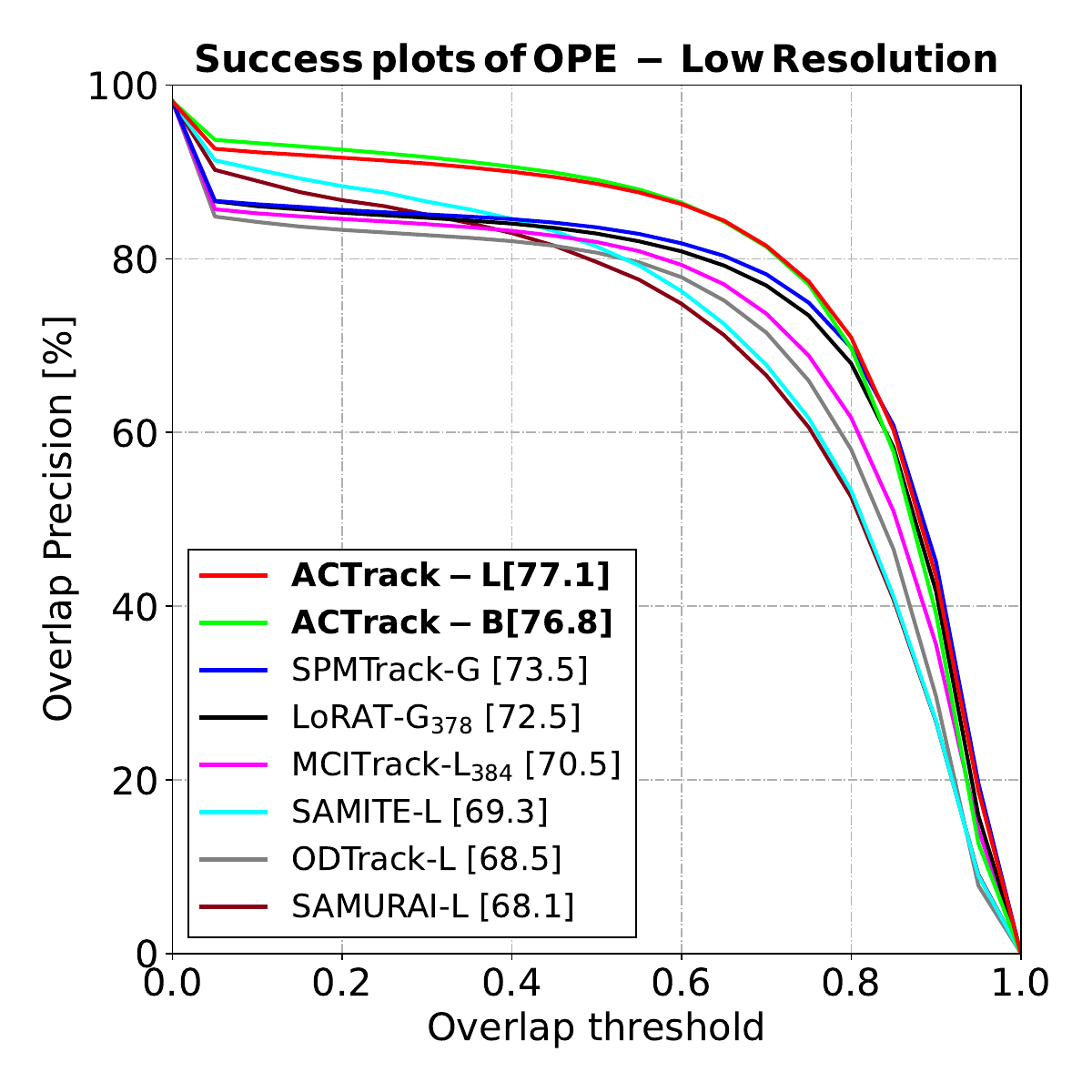}
\end{minipage}\hfill
\begin{minipage}{0.193\textwidth}
\centering
\includegraphics[width=\linewidth]{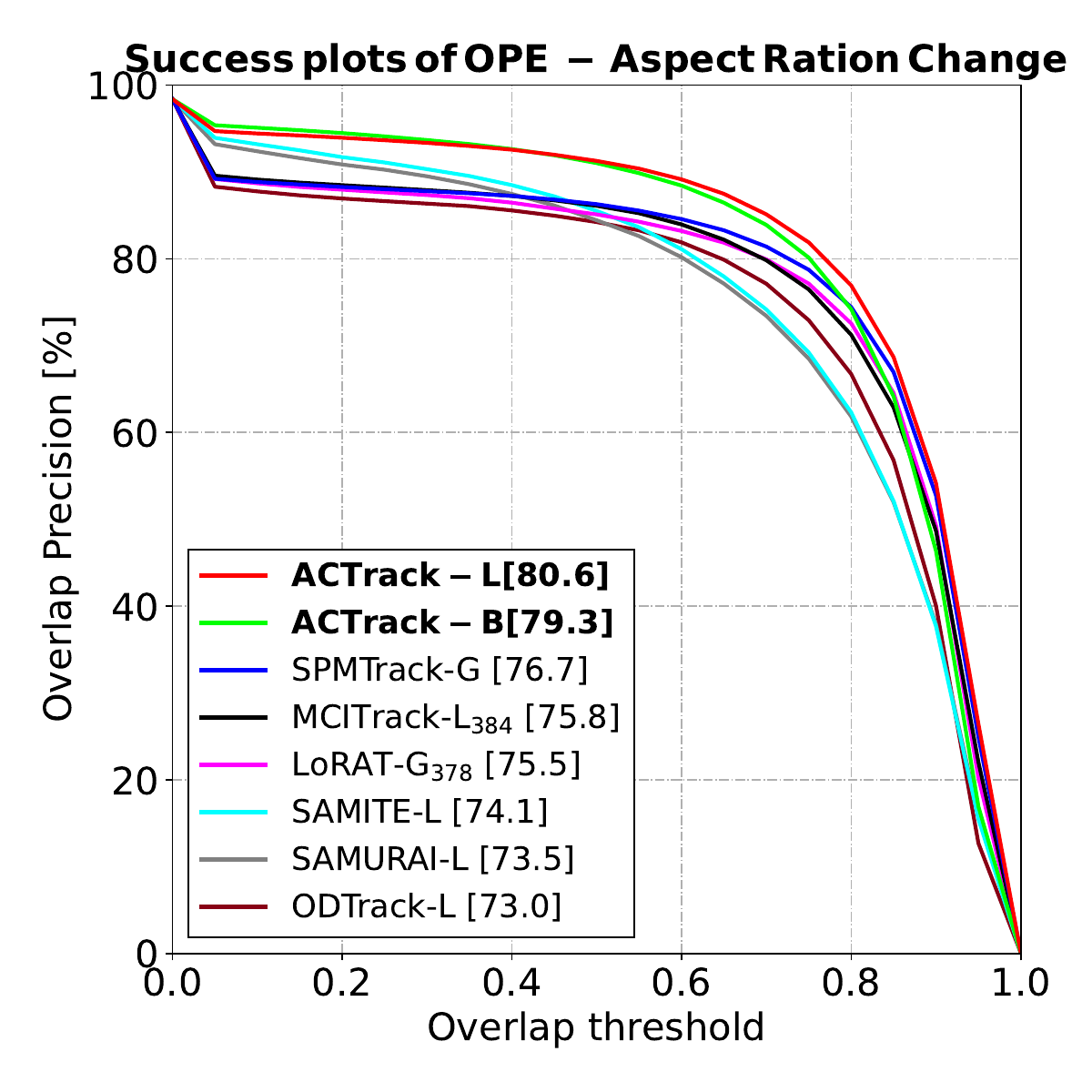}
\end{minipage}\hfill
\begin{minipage}{0.193\textwidth}
\centering
\includegraphics[width=\linewidth]{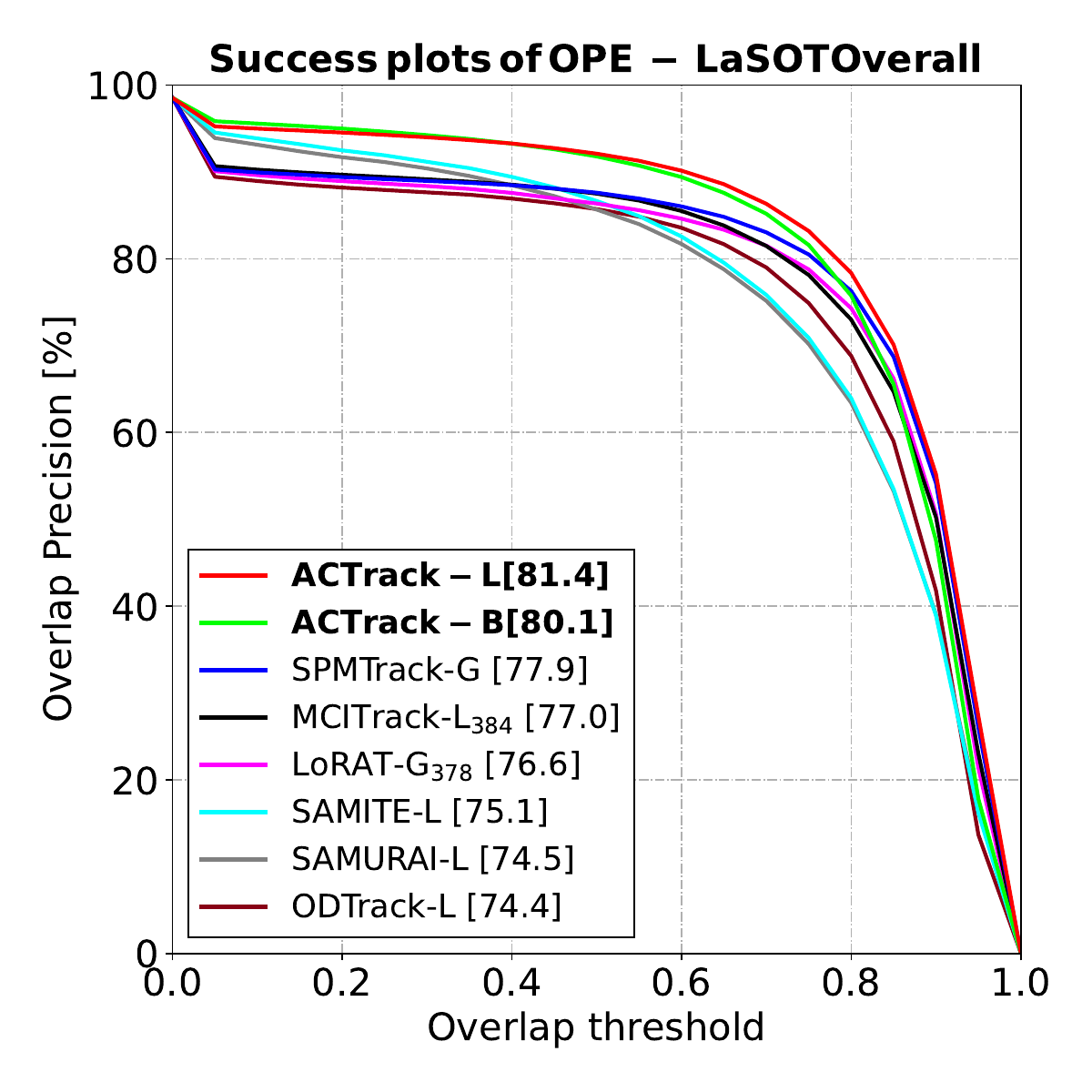}
\end{minipage}
\caption{Comparisons of our proposed ACTrack-B and ACTrack-L with other excellent trackers in the success curve on LaSOT \emph{test} split, which includes fourteen challenging scenarios such as Low Resolution, Motion Blur, Scale Variation, etc. We also provide the comparison of the success curve across the entire LaSOT \emph{test} split.}
\label{fig:comparision of lasot dataset}
\vspace{-1ex}
\end{figure*}

\begin{figure*}[!t]
  \centering
    \subfloat[Qualitative results of three methods when the targets is moving rapidly.]{\includegraphics[width=\textwidth]{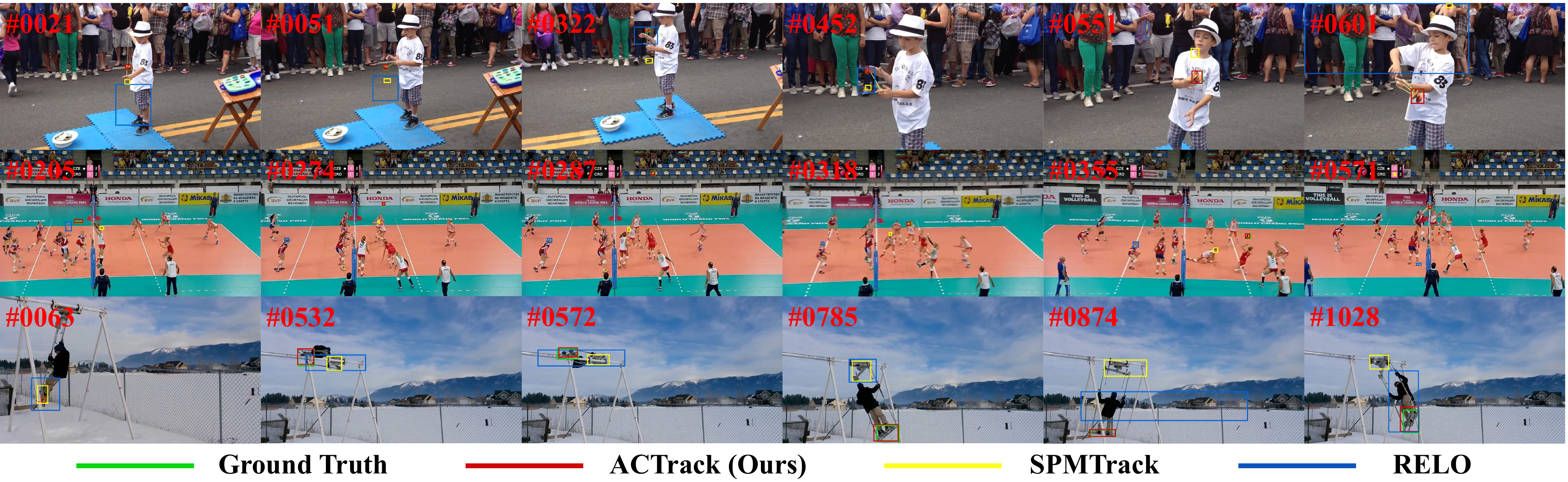}\label{fig-vis-2}} \\
    \subfloat[Qualitative results of three methods when the targets becomes blurred in motion.]{\includegraphics[width=\textwidth]{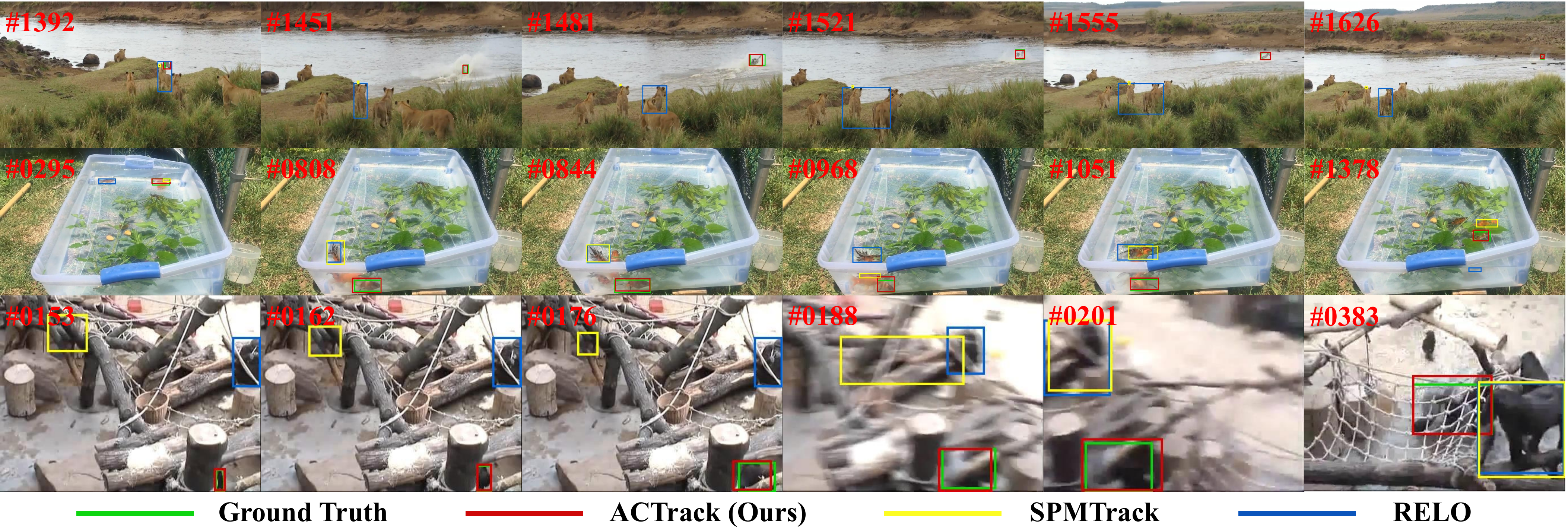}\label{fig-vis-3}} \\
    \subfloat[ Qualitative results of three methods when the targets have large occlusions.]{\includegraphics[width=\textwidth]{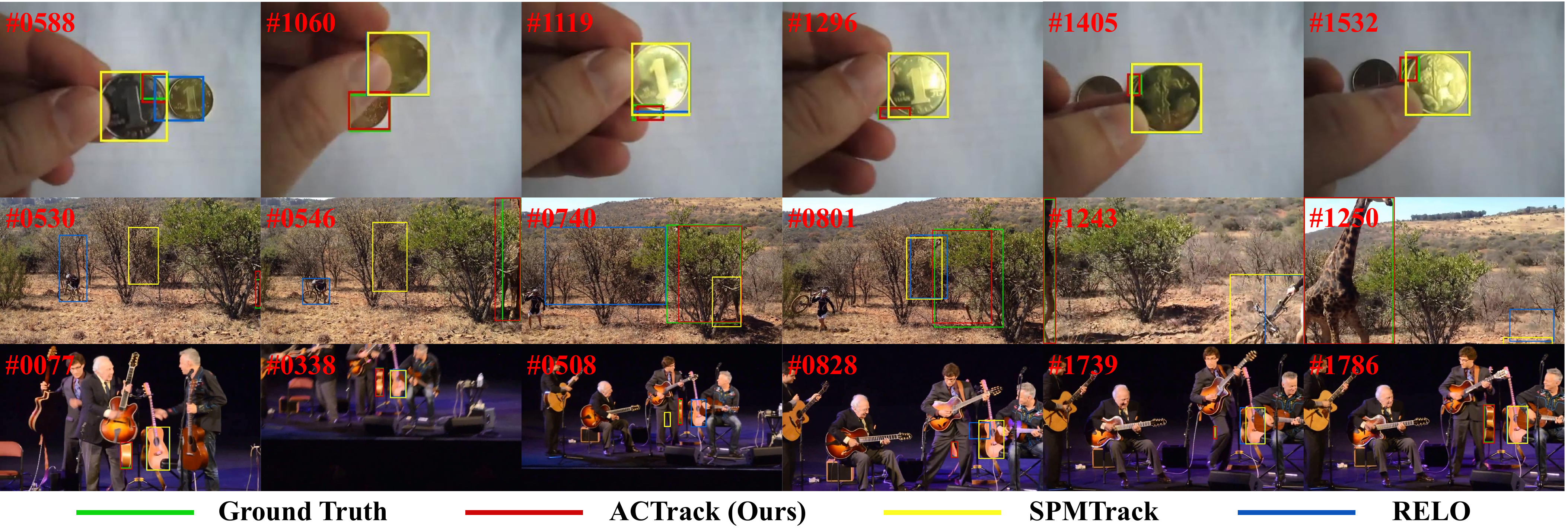}\label{fig-vis-4}} \\
  \caption{ This figure presents a visual comparison among our proposed ACTrack-B, SPMTrack~\cite{Cai_2025_CVPR_SPMTrack} and RELO~\cite{chen2026reloreinforcementlearninglocalize} in the challenges of target undergoes sudden movements, motion blur, and severe occlusion. It demonstrates that our method achieves more effective and accurate tracking in the aforementioned challenging scenarios. Zoom in for better view.}
    \label{fig:qualitative}
\end{figure*}

\begin{figure*}[!t]
  \centering
    \vspace{3ex}
    \subfloat[Qualitative results of three methods on RGB-Depth tasks.]{\includegraphics[width=\textwidth]{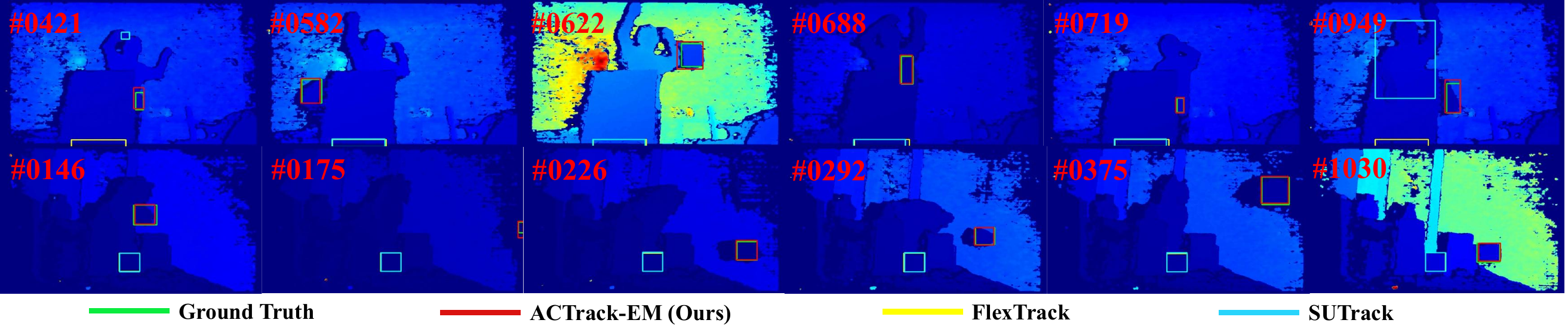}\label{fig-vis-1}} \\
    \subfloat[ Qualitative results of three methods on RGB-Event tasks. ]
    {\includegraphics[width=\textwidth]{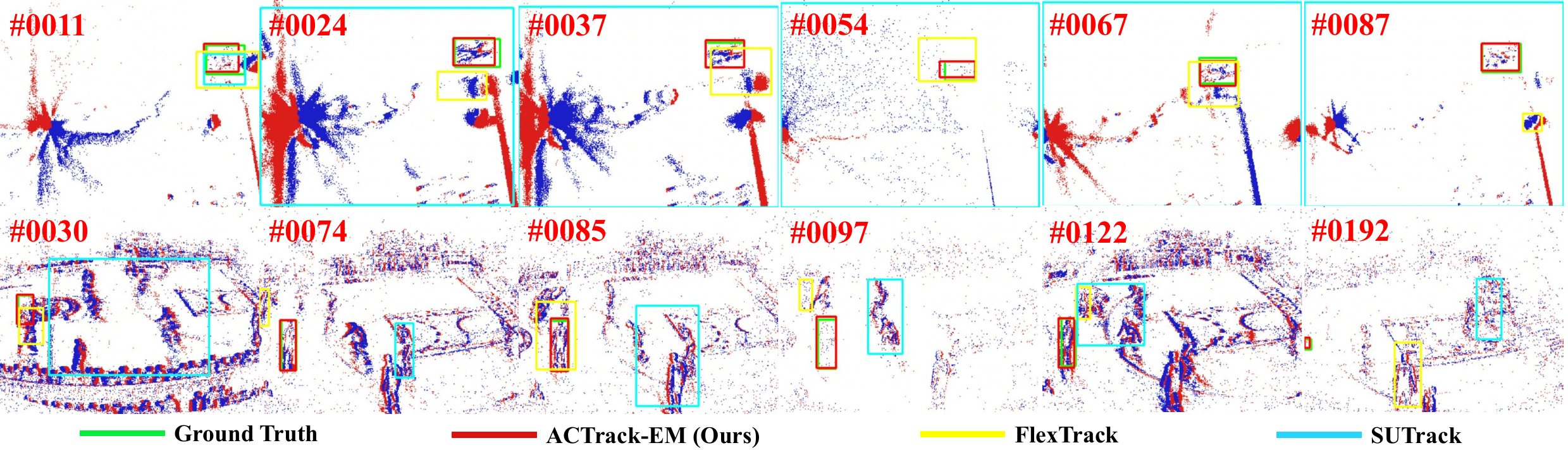}\label{fig-vis-2}} \\
    \subfloat[ Qualitative results of three methods on RGB-Thermal tasks.]{\includegraphics[width=\textwidth]{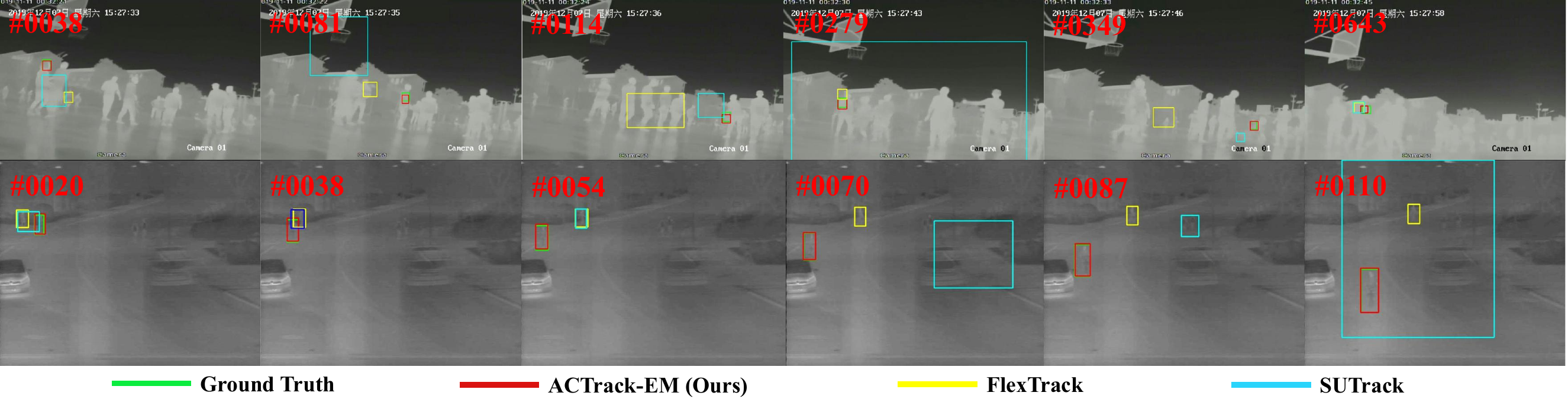}\label{fig-vis-3}} \\
    \vspace{-1ex}
  \caption{ This figure presents a visual comparison among our proposed ACTrack-EM,  FlexTrack \cite{tan2025you_flextrack} and SUTrack \cite{sutrack} in the challenges of different modalities including RGB-Depth, RGB-Event and RGB-Thermal tasks. It demonstrates that our method achieves more effective and accurate tracking in the aforementioned challenging scenarios. Zoom in for better view.}
  \vspace{-1ex}
    \label{fig:quantitative2}
\end{figure*}

\begin{table}
\centering
\caption{Tracking capability of individual tools and the overall ACTrack-B$_{224}$ framework on the LaSOT \emph{test} split.}
\label{tab:individual_tool_capability}
\footnotesize
\setlength{\tabcolsep}{6pt}
\begin{tabular*}{\columnwidth}{@{\extracolsep{\fill}}l|ccc}
\toprule
\textbf{Method} & \textbf{AUC} & $\mathbf{P_{norm}}$ & \textbf{P} \\
\midrule
Instance Matching Tool & 75.3 & 85.6 & 83.3 \\
SAM3 Motion Tool & 76.3 & 82.8 & 80.9   \\
ACTrack-B$_{224}$ & \textbf{80.1} & \textbf{90.3} & \textbf{87.8} \\
\bottomrule
\end{tabular*}
\end{table}

\begin{table}
\centering
\caption{Effect of progressively activating additional tools on top of the Instance Matching Tool. All variants share MCITrack-B$_{224}$ as the Instance Matching Tool and are evaluated on the LaSOT \emph{test} split.}
\label{tab:tool_coordination}
\footnotesize
\setlength{\tabcolsep}{3pt}
\begin{tabular*}{\columnwidth}{@{\extracolsep{\fill}}l|ccc}
\toprule
\textbf{Configuration} & \textbf{AUC} & $\mathbf{P_{norm}}$ & \textbf{P} \\
\midrule
Matching only & 75.3 & 85.6 & 83.3 \\
 + Motion Region & 78.9 & 88.7 & 86.2 \\
 + Perception Conflict & 79.6 & 89.6 & 87.0  \\
 + Reprompt (ACTrack-B$_{224}$) & \textbf{80.1} & \textbf{90.3} & \textbf{87.8} \\
\bottomrule
\end{tabular*}
\end{table}

\begin{table}
\centering
\caption{Effect of the search-region cropping factor $\rho$ on different datasets in terms of AUC (F-score for DepthTrack). RGB results use ACTrack-B$_{224}$, and the remaining modalities use ACTrack-EM.}
\label{tab:search_region_cropping_factor}
\footnotesize
\setlength{\tabcolsep}{3pt}
\begin{tabular*}{\columnwidth}{@{\extracolsep{\fill}}l|cccc}
\toprule
\multirow{2}{*}{\textbf{Dataset (Modality)}} & \multicolumn{4}{c}{\textbf{Cropping factor }$\rho$} \\
\cmidrule(lr){2-5}
 & $1.5$ & $2.0$ & $2.5$ & $4.0$ \\
\midrule
LaSOT (RGB)        & 79.8        & 79.6 & \textbf{80.1} & 79.6 \\
TNL2K (RGB-L)      & 77.2        & 77.1 & \textbf{77.2} & 77.2 \\
VisEvent (RGB-E)   & \textbf{77.8} & 77.7 &  77.6       & 77.7 \\
LasHeR (RGB-T)     & \textbf{63.7} & 63.5 &  63.1   & 63.4 \\
DepthTrack (RGB-D) & \textbf{74.9} & 73.5 &  73.3   & 73.7 \\
\bottomrule
\end{tabular*}
\end{table}

\begin{table}
\centering
\caption{Effect of the number of consecutive conflicting frames $K$ that triggers the VLM Reprompt Tool, reported in terms of AUC (F-score for DepthTrack). 
``Calls'' is the average number of VLM invocations per sequence. RGB results use ACTrack-B$_{224}$, and the remaining modalities use ACTrack-EM.}
\label{tab:vlm_trigger_interval}
\footnotesize
\setlength{\tabcolsep}{4pt}
\begin{tabular*}{\columnwidth}{@{\extracolsep{\fill}}l|ccc|c}
\toprule
\multirow{2}{*}{\textbf{Dataset (Modality)}} & \multicolumn{3}{c|}{\textbf{AUC at trigger interval }$K$} & \textbf{Calls} \\
\cmidrule(lr){2-4}
 & \textbf{$5$} & $10$ & $20$ & ($K=5$) \\
\midrule
LaSOT (RGB)        & \textbf{80.1} & 79.5 & 79.5 & 1.3 \\
TNL2K (RGB-L)      & \textbf{77.2} & 77.2 & 77.1 & 0.7 \\
VisEvent (RGB-E)   & \textbf{77.8} & 77.8 & 77.8 & 0.3 \\
LasHeR (RGB-T)     & \textbf{63.7} & 63.3 & 63.0 & 2.4 \\
DepthTrack (RGB-D)     & \textbf{74.9} & 74.0 & 72.8 & 2.9 \\
\bottomrule
\end{tabular*}
\end{table}

\begin{table}[!h]
\centering
\caption{A performance comparison of existing trackers and their integration with our proposed ACTrack framework on the LaSOT \emph{test} set.}
\label{tab:generality}
\footnotesize
\setlength{\tabcolsep}{4pt}
\begin{tabular*}{\columnwidth}{@{\extracolsep{\fill}}l|ccc}
\toprule
\textbf{Method} & \textbf{AUC} & $\mathbf{P_{norm}}$ & \textbf{P} \\
\midrule
ODTrack-L$_{384}$~\cite{ODTrack}        & $74.0$ & $84.2$ & $82.3$ \\
ODTrack-L$_{384}$ w/ ACTrack            & $\mathbf{78.1}$ & $\mathbf{87.9}$ & $\mathbf{84.9}$ \\
\midrule
MCITrack-L$_{384}$~\cite{kang2025exploring_mcitrack} & $76.6$ & $86.1$ & $85.0$ \\
MCITrack-L$_{384}$ w/ ACTrack           & $\mathbf{81.4}$ & $\mathbf{90.8}$ & $\mathbf{89.4}$ \\
\bottomrule
\end{tabular*}
\end{table}

\begin{table*}
\centering
\caption{Ablation on SAM3 motion configurations on the LaSOT \emph{test} split.
 Each row is evaluated in two roles: as a standalone tracker (left block), and as the Motion Tool that provides the search region of the Instance Matching Tool in ACTrack-B$_{224}$ (right block).}
\label{tab:sam3_motion_ablation}
\footnotesize
\setlength{\tabcolsep}{4pt}
\begin{tabular*}{\textwidth}{@{\extracolsep{\fill}}l|ccc|ccc}
\toprule
\multirow{2}{*}{\textbf{SAM3 Configuration}} & \multicolumn{3}{c|}{\textbf{SAM3 alone}} & \multicolumn{3}{c}{\textbf{SAM3 + Instance Matching Tool}} \\
\cmidrule(lr){2-4} \cmidrule(lr){5-7}
 & \textbf{AUC} & $\mathbf{P_{norm}}$ & \textbf{P} & \textbf{AUC} & $\mathbf{P_{norm}}$ & \textbf{P} \\
\midrule
Vanilla SAM3 & 75.1 & 82.0 & 79.9 & 78.4 & 88.1 & 85.7 \\
\texttt{w/} KF ($W_{\mathrm{m}}=0.05$, $P_\mathrm{obj}=0.5$) & 75.7 & 82.5 & 80.6 & 78.8 & 88.6 & 86.0  \\
\texttt{w/} KF ($W_{\mathrm{m}}=0.15$, $P_\mathrm{obj}=0.5$) & 76.3 & 82.8 & 80.9 & \textbf{78.9} & \textbf{88.7} & \textbf{86.2} \\
\texttt{w/} KF ($W_{\mathrm{m}}=0.30$, $P_\mathrm{obj}=0.5$) & 76.0 & 82.7 & 80.7 & 78.9 & 88.6 & 86.2 \\
\texttt{w/} KF ($W_{\mathrm{m}}=0.15$, $P_\mathrm{obj}=0.269$) & 76.5 & 83.2 & 81.1 & 78.7 & 88.4 & 85.9 \\
\texttt{w/} KF and Memory Selection~\cite{xu2025samite} & \textbf{76.9} & \textbf{83.5} & \textbf{81.4} & 78.8 & 88.5 & 86.1 \\
\bottomrule
\end{tabular*}
\end{table*}

\begin{table*}
\centering
\caption{Ablation of the fine-tuning design of the standalone Instance Matching Tool used in ACTrack-EM, in terms of AUC on LaSOT, VisEvent, LasHeR, DepthTrack, and TNL2K. ``PE'' denotes patch embedding. The ``Param'' column reports the total parameters of the \emph{standalone Instance Matching Tool}, with the parameters of the patch-embedding module shown in parentheses.}
\label{tab:em_finetune_design}
\footnotesize
\setlength{\tabcolsep}{6pt}
\begin{tabular*}{\textwidth}{@{\extracolsep{\fill}}l|cccccc}
\toprule
\textbf{Configuration} & \textbf{LaSOT} & \textbf{VisEvent} & \textbf{LasHeR} & \textbf{DepthTrack} & \textbf{TNL2K} & \textbf{Param (M)} \\
\midrule
Shared PE                       & \textbf{76.2} & 68.7 & 59.8 & 65.9 & 64.2 & 495.6 (1.2) \\
+ Modality-specific Unified PE  & 75.6 & 69.5 & 60.9 & 66.4 & 64.5 & 496.8 (2.4) \\
+ Residual fusion (ours)        & 76.0 & \textbf{69.8} & \textbf{61.3} & \textbf{66.9} & \textbf{64.8} & 498.9 (4.5) \\
\bottomrule
\end{tabular*}
\end{table*}




\subsection{More Detailed Results in Different Attribute Scenes}

LaSOT~\cite{fan2019lasot} is well-known for featuring a diverse range of challenging tracking scenarios, therefore, 
in Fig.~\ref{fig:comparision of lasot dataset}, we provide a more detailed comparison of our proposed ACTrack-B$_{224}$ and ACTrack-L$_{384}$ with other current excellent trackers SPMTrack-G~\cite{Cai_2025_CVPR_SPMTrack}, LoRAT-G$_{378}$~\cite{LoRAT}, MCITrack-L$_{384}$, SAMITE-L~\cite{xu2025samite}, SAMURAI-L~\cite{yang2026samurai} and ODTrack-L~\cite{ODTrack} across various challenging scenario subsets in LaSOT \cite{fan2019lasot}.
Fig.~\ref{fig:comparision of lasot dataset} presents detailed success curves and AUC scores across individual subsets, along with the success curve on the entire LaSOT \emph{test} split. The results demonstrate that our ACTrack significantly outperforms these RGB-based trackers both overall and across the vast majority of subsets.

\subsection{Ablation Studies}

\noindent \textbf{Tracking Capability of Individual Tools.}
Within the overall ACTrack framework, 
both the Instance Matching Tool and the SAM3-based Motion Tool possess tracking capability. 
In Table~\ref{tab:individual_tool_capability}, we compare the tracking capability of each component within ACTrack-B, as well as the overall tracking capability. The results show that the overall tracking capability of ACTrack significantly outperforms that of each individual tool, demonstrating the effectiveness of the overall framework.

\noindent \textbf{Effect of Tool Coordination.}
We isolate the contribution of each tool by starting from MCITrack-B$_{224}$ used alone as the Instance Matching Tool and incrementally activating the remaining tools. The corresponding LaSOT scores are reported in Table~\ref{tab:tool_coordination}. Re-anchoring the search region with the Motion Tool already yields a clear gain over the standalone tracker; the instance-conflict detection of the Perception Tool brings a further improvement; and enabling the VLM Reprompt Tool produces the best result, corresponding to the full ACTrack configuration.

\noindent \textbf{The Impact of Search Region Cropping Factor based on the Motion Tool.} 
In ACTrack, the search region of the Instance Matching Tool is cropped based on the Motion Tool. In Table~\ref{tab:search_region_cropping_factor}, we compare the effect of different search region cropping factors for ACTrack-B$_{224}$ on the RGB dataset and ACTrack-EM across datasets in other modalities.
As shown in Table~\ref{tab:search_region_cropping_factor}, different search region cropping factors do not substantially affect the overall performance. On the RGB dataset, a cropping factor of $2.5$ yields better performance, whereas on datasets of other modalities, a cropping factor of $1.5$ yields better performance.

\noindent \textbf{Triggering Interval of the VLM Reprompt Tool.}
In ACTrack, the VLM Reprompt Tool is invoked once the predictions of the Motion Tool and the Instance Matching Tool remain inconsistent for a sustained span of frames; the VLM then arbitrates between the two candidates and corrects the Motion Tool whenever it has drifted. 
In Table~\ref{tab:vlm_trigger_interval}, we study how the number of consecutive conflicting frames $K$ required to trigger the VLM Reprompt Tool affects the overall performance. 
The results indicate that triggering on five consecutive conflicting frames achieves the best accuracy, while the number of invocations stays sufficiently small to keep the impact on overall inference speed negligible.
In practice, the VLM is called only a few times per sequence, and each VLM call takes about 2.1 seconds on average, so this sparse reprompting mechanism has little impact on the overall efficiency.

\noindent \textbf{Generality of the ACTrack Framework.}
The variants reported throughout this paper instantiate the Instance Matching Tool with MCITrack-B$_{224}$, MCITrack-L$_{384}$, and the parameter-efficient fine-tuned tracker in ACTrack-EM. 
To further verify that ACTrack remains effective with different underlying trackers, Table~\ref{tab:generality} compares additional choices of Instance Matching Tools within the same coordination framework. 
The results show that ACTrack consistently delivers substantial gains regardless of the tracker used as the Instance Matching Tool, while a stronger underlying tracker naturally translates into stronger overall performance.

\noindent \textbf{Improvements on the SAM3-based Motion Tool.}
ACTrack augments the original SAM3 video tracker with a Kalman filter for motion modeling. 
In Table \ref{tab:sam3_motion_ablation}, we compare the tracking capability of vanilla SAM3 with that of the Kalman-augmented variant. 
We further study the influence of the motion-prior weight and of incorporating a SAMITE-style memory selection mechanism~\cite{xu2025samite}. 
The results in Table \ref{tab:sam3_motion_ablation} show that the strongest standalone SAM3 configuration is not necessarily the best Motion Tool configuration inside ACTrack.
For example, the motion-prior weight adopted in this paper is not the optimum for standalone SAM3 tracking, 
and the SAMITE-style memory selection leads to noticeable improvements when SAM3 is used in isolation but it does not bring the best result after the Instance Matching Tool is introduced. 
This is because both modifications aim to mitigate the instance confusion and memory contamination issues of SAM3 discussed earlier, whereas in ACTrack the cooperation with the remaining tools already alleviates these issues. 
Table \ref{tab:sam3_motion_ablation} also reports the performance obtained when search regions produced by different SAM3 motion-tool configurations are provided to the Instance Matching Tool, the resulting tracker not only substantially outperforms the strengthened SAM3 baselines, but also confirms that the configuration adopted in this paper achieves the best performance overall.

\noindent \textbf{Ablation of the Fine-tuned Instance Matching Tool in ACTrack-EM.}
In ACTrack-EM, the Instance Matching Tool is obtained by parameter-efficient fine-tuning of the SAM3 image encoder. 
To separate its own design choices from the full ACTrack coordination policy, we evaluate this tool alone in Table~\ref{tab:em_finetune_design}.
The shared patch embedding remains competitive and gives the highest LaSOT score.
However, the modality-specific unified patch embedding improves all remaining benchmarks while adding only 1.2M patch-embedding parameters.
When each modality is assigned its corresponding patch embedding, the training gradients dedicated to RGB discrimination are diluted accordingly, which may slightly degrade the performance of the pure RGB modality.
Adding the residual fusion connection further improves every benchmark over the modality-specific unified patch embedding and therefore gives a better overall multimodal trade-off with a small parameter increase.
Table~\ref{tab:em_aux_template} further examines the dynamic auxiliary templates.
Using two auxiliary templates gives the best result among the tested choices, and updating them from high-confidence search frames every $\lfloor n/5 \rfloor$ frames is the most effective setting in this grid.

\begin{table}[!h]
\centering
\caption{Ablation of the auxiliary-template settings of the standalone Instance Matching Tool used in ACTrack-EM on the LaSOT \emph{test} split. $n$ denotes the number of frames already tracked.}
\label{tab:em_aux_template}
\footnotesize
\setlength{\tabcolsep}{4pt}
\begin{tabular*}{\columnwidth}{@{\extracolsep{\fill}}l|c|ccc}
\toprule
\textbf{Factor} & \textbf{Setting} & \textbf{AUC} & $\mathbf{P_{norm}}$ & \textbf{P} \\
\midrule
\multirow{3}{*}{\# Aux templates} & $0$ & 75.0 & 82.9 & 81.3 \\
                                  & $1$ & 75.4 & 83.6 & 81.9 \\
                                  & \textbf{$2$} & \textbf{76.0} & \textbf{84.3} & \textbf{82.7} \\
\midrule
\multirow{3}{*}{Update threshold} & $0.5$ & 74.5 & 82.5 & 80.7 \\
                                  & $0.75$ & 75.2 & 83.4 & 81.5 \\
                                  & \textbf{$0.9$} & \textbf{76.0} & \textbf{84.3} & \textbf{82.7} \\
\midrule
\multirow{3}{*}{Update interval}  & $\lfloor n/10 \rfloor$ & 75.4 & 83.7 & 81.8 \\
                                  & \textbf{$\lfloor n/5 \rfloor$} & \textbf{76.0} & \textbf{84.3} & \textbf{82.7} \\
                                  & $\lfloor n/2 \rfloor$ & 75.6 & 83.8 & 82.0 \\
\bottomrule
\end{tabular*}
\end{table}

\subsection{Qualitative Study}

\subsubsection{Results On RGB Visual Tracking}

We further analyze representative sequences to understand when tool coordination helps. We provide the detail visualization results in Fig.~\ref{fig:qualitative}.
All videos are from the \emph{test} split of LaSOT. We compare our proposed ACTrack with SPMTrack~\cite{Cai_2025_CVPR_SPMTrack} and RELO~\cite{chen2026reloreinforcementlearninglocalize} in terms of performance when the target undergoes sudden movement, motion blur, and severe occlusion. All the selected videos are challenging, as described below:
\begin{itemize}
    \item Fig.~\ref{fig:qualitative}(a) demonstrates the tracking results of three methods when the target suffers from sudden movement.
    \item Fig.~\ref{fig:qualitative}(b) demonstrates the tracking results of three methods when the targets suffers from motion blur.
    \item Fig.~\ref{fig:qualitative}(c) demonstrates the tracking results of three methods when the target suffers from severe occlusion.
\end{itemize}

We observe that in fast motion scenarios (as shown in Fig.~\ref{fig:qualitative}(a)),
previous trackers commonly drift, whereas our method can even successfully track a fast-moving volleyball. Moreover, in scenarios involving partial occlusion combined with motion blur, as well as scenarios with nearly complete occlusion, our method demonstrates tracking capability that far surpasses that of previous methods.

\subsubsection{Results on Multi-Modal Visual Tracking}

We further provide visualized comparisons of our proposed ACTrack-EM against other excellent trackers SUTrack \cite{sutrack}, and FlexTrack \cite{tan2025you_flextrack} across other modalities including RGB-Depth in Fig.~\ref{fig:quantitative2}(a), RGB-Event in Fig.~\ref{fig:quantitative2}(b) and RGB-Thermal in Fig.~\ref{fig:quantitative2}(c).
Our ACTrack-EM consistently  exhibit superior performance on these modalities. 


\section{Conclusion}
\label{sec:conclusion}

We presented ACTrack, an agentic tracking framework that coordinates heterogeneous visual models as tools. The Instance Matching Tool provides efficient box-level localization, the SAM3 Motion and Perception Tools provide mask-derived motion priors and instance-level conflict evidence, and the VLM Reprompt Tool is invoked only under persistent disagreement for sparse identity arbitration and prompt renewal. This separates routine localization from motion propagation, instance perception, and semantic reasoning, turning visual tracking into a closed-loop decision process rather than a single-model matching problem.

This work also argues for a broader view of visual tracking. Matching-based trackers remain important, but simply making them larger or more specialized should not be the only path forward. Visual foundation models and multimodal large models already provide capabilities that conventional trackers lack, including promptable segmentation, object-level memory, and semantic identity reasoning. ACTrack is an initial demonstration that these capabilities can be organized into a practical tracking pipeline. We hope it encourages further exploration of foundation-model and multimodal-model tools for future tracking systems.

\section*{Data Availability Statements}

The datasets used in this manuscript are all deposited in publicly available repositories. LaSOT and LaSOT$_\mathrm{ext}$~\cite{fan2019lasot} can be obtained from \url{https://github.com/HengLan/LaSOT_Evaluation_Toolkit}. TrackingNet~\cite{muller2018trackingnet} can be obtained from \url{https://github.com/SilvioGiancola/TrackingNet-devkit}. GOT-10k~\cite{huang2019got} can be obtained from \url{https://github.com/got-10k/toolkit}. TNL2K~\cite{Wang_2021_CVPR_TNL2k} can be obtained from \url{https://github.com/wangxiao5791509/TNL2K_evaluation_toolkit}. OTB2015~\cite{otb2015} can be obtained from \url{http://cvlab.hanyang.ac.kr/tracker_benchmark/}. UAV123~\cite{mueller2016benchmark} can be obtained from \url{https://ivul.kaust.edu.sa/benchmark-and-simulator-uav-tracking-dataset}. NfS~\cite{Galoogahi_2017_ICCV_nfs} can be obtained from \url{http://ci2cv.net/nfs/index.html}. VastTrack~\cite{NEURIPS2024_ec17a52e_vasttrack} can be obtained from \url{https://github.com/HengLan/VastTrack}. LasHeR~\cite{9640453_lasher} can be obtained from \url{https://github.com/BUGPLEASEOUT/LasHeR}. VisEvent~\cite{10284004_visevent} can be obtained from \url{https://github.com/wangxiao5791509/VisEvent_SOT_Benchmark}. DepthTrack~\cite{Yan_2021_ICCV_depthtrack} can be obtained from \url{https://github.com/xiaozai/DeT}. COCO~\cite{lin2014microsoft} can be obtained from \url{https://cocodataset.org/}.



\bibliography{sn-bibliography}

\end{document}